\documentclass[10pt,twocolumn,letterpaper]{article}

\usepackage{cvpr}              
\usepackage{bm}
\usepackage{bbm}
\usepackage{makecell}
\usepackage{algorithm}
\usepackage{algpseudocode}
\usepackage[inkscapelatex=false]{svg}

\definecolor{cvprblue}{rgb}{0.21,0.49,0.74}
\usepackage[pagebackref,breaklinks,colorlinks,allcolors=cvprblue]{hyperref}

\def\paperID{8067} 
\def\confName{CVPR}
\def\confYear{2025}

\newcommand\blfootnote[1]{%
  \begingroup
  \vspace{-0.25cm}
  \renewcommand\thefootnote{}\footnote{#1}%
  \addtocounter{footnote}{-1}%
  \endgroup
}

\title{Prof. Robot: Differentiable Robot Rendering Without Static and Self-Collisions}

\author{Quanyuan Ruan, Jiabao Lei, Wenhao Yuan, Yanglin Zhang, Dekun Lu, Guiliang Liu, Kui Jia}
\author{
Quanyuan Ruan$^1$\footnotemark[2]
\and
Jiabao Lei$^2$\footnotemark[2]
\and
Wenhao Yuan$^1$        
\and
Yanglin Zhang$^2$
\and
Dekun Lu$^1$
\and
Guiliang Liu$^2$\footnotemark[1]
\and
Kui Jia$^2$\footnotemark[1]
\and
$^1$South China University of Technology \\
$^2$School of Data Science, The Chinese University of Hong Kong, Shenzhen \\
}

\begin{document}


\twocolumn[{%
\renewcommand\twocolumn[1][]{#1}%
\maketitle
\begin{center}
    \vspace{-0.5cm}
    \centering
    \captionsetup{type=figure}
    \includegraphics[width=\textwidth]{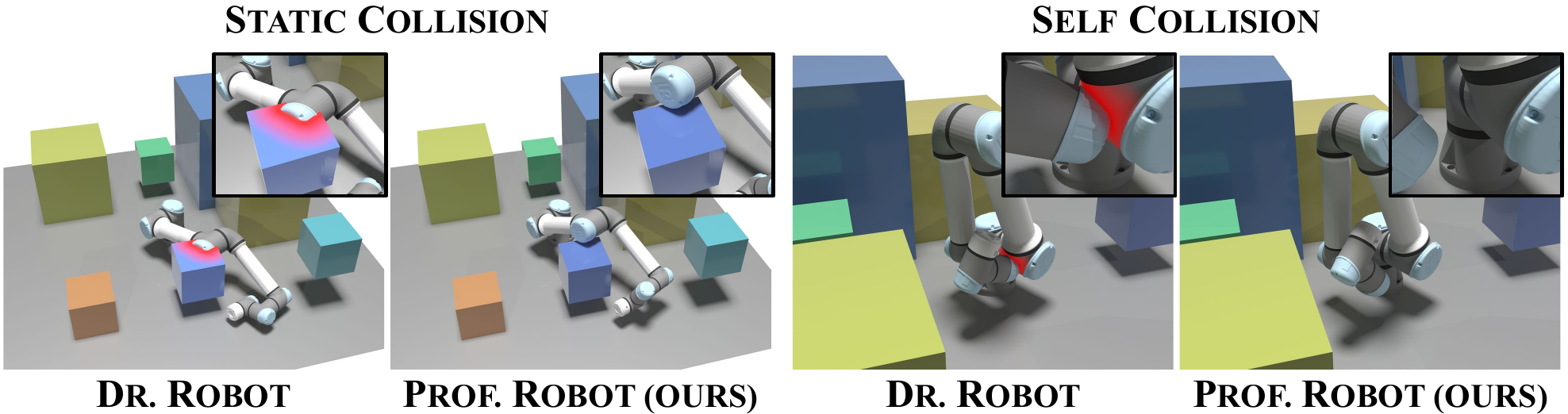}
    \captionof{figure}{\textbf{Our method improves upon Dr.Robot~\cite{liu2024differentiable} by additionally enabling differentiable avoidance of static and self-collisions.} By learning and integrating a gradient-consistent pose classifier into a differentiable rendering pipeline, the generated pose trajectories are free from physical collisions. The objective is to penalize high collision probabilities during optimization. On the left, our method can push robotic arms away from intersecting objects. On the right, it can also avoid self-collisions. Images are rendered using Blender with meshes.}
\end{center}%
}]

\blfootnote{\textsuperscript{\dag} Equal contribution.}
\blfootnote{\textsuperscript{*} Corresponding authors.}
\blfootnote{\hspace{0.65em}Project: \url{https://qrcat.github.io/prof-robot/}}

\begin{abstract}
Differentiable rendering has gained significant attention in the field of robotics, with differentiable robot rendering emerging as an effective paradigm for learning robotic actions from image-space supervision. However, the lack of physical world perception in this approach may lead to potential collisions during action optimization. In this work, we introduce a novel improvement on previous efforts by incorporating physical awareness of collisions through the learning of a neural robotic collision classifier. This enables the optimization of actions that avoid collisions with static, non-interactable environments as well as the robot itself. To facilitate effective gradient optimization with the classifier, we identify the underlying issue and propose leveraging Eikonal regularization to ensure consistent gradients for optimization. Our solution can be seamlessly integrated into existing differentiable robot rendering frameworks, utilizing gradients for optimization and providing a foundation for future applications of differentiable rendering in robotics with improved reliability of interactions with the physical world. Both qualitative and quantitative experiments demonstrate the necessity and effectiveness of our method compared to previous solutions.
\end{abstract}
\section{Introduction}

Conventional methods for representing a robot involve partitioning it into many rigid body parts and then articulating them according to structural kinematic constraints. 
However, they only model the robots geometrically and do not possess a realistic appearance for rendering; therefore, they are less applicable for integration into a differentiable pipeline involving images, which is currently a popular approach in computer vision.
A recent approach~\cite{liu2024differentiable} has been proposed to model robots' appearance in a differentiable manner, facilitating the integration of robots into a differentiable rendering solution. 
It allows for backpropagating the appearance supervision from images to the control space, thereby enabling the learning of plausible robot movements.
Their method is straightforward. They leverage Gaussian splatting~\cite{kerbl3Dgaussians} for efficient differentiable rendering, where each Gaussian point is associated with the robotic arms using implicit LBS~\cite{loper2023smpl, liu2024differentiable}, and the arms are generated by the control parameters through forward kinematics.

While promising, their method clearly lacks awareness of how to avoid physical collisions with a static environment that does not dynamically interact with the robot, as well as collisions with the robot itself. 
When optimized using gradient information, it is likely to produce invalid robotic parameters that lead to intersections. 
This problem is particularly severe in real scenarios, where minor collisions damage the robot's mechanical structure. Due to the typically high cost of robots, which can result in significant financial losses.
To address this issue and ensure that the differentiable rendering pipeline produces valid robotic parameters that prevent collisions, we need to guide the robot to evade collision scenarios during gradient-guided optimization. 
Intuitively, we employ a similar approach to that in \cite{tiwari2022pose} to achieve collision-free optimization, with some critical adaptations and improvements.

Our idea is simple. We observe that some robotic poses are free from collisions, while others are not. Therefore, we can classify these poses based on whether they lead to a collision. We can leverage a discriminative model to classify these poses. The classifier $f$ is a mapping \(\textrm{SO}(2)^K \rightarrow [0, 1]\) that maps a pose \(\bm{\theta} \in \textrm{SO}(2)^K\) to a decision \(c \in [0, 1]\), where a value of $c>0.5$ indicates a collision. A rotation occurs in \(\textrm{SO}(2)\) because the motor of a robotic joint typically possesses only one degree of freedom for rotation. Finally, we need to integrate the model into a differentiable pipeline. During optimization, to ensure that the robotic pose remains in the collision-free zone, we must push the pose \(\bm{\theta}\) toward the collision-free regions $\{\bm{\theta} \in \textrm{SO}(2)^K\mid f(\bm{\theta}) \le 0.5\}$ that correspond to a classification result of $c \le 0.5$. Compared to \cite{tiwari2022pose}, we should note that the biggest benefit of using classification rather than fitting a distance field is that classification does not require computing manifold distances and only needs to provide classification labels, corresponding to either ``collided'' or ``not collided.'' Computing labels is expensive and may lack accuracy, as the normally generated samples are often too sparse to accurately compute the manifold distances in such a high-dimensional space due to the curse of dimensionality.

Unfortunately, the direct application of the above idea does not yield the desired results due to the variability of the gradients obtained after sheer classification learning. 
The update direction of the gradients may not be consistent, and a later update may offset the previous one. We will mathematically analyze the reasons for this phenomenon in Section~\ref{sec:eikonal_regularizer_for_consistent_gradients}. 
To address this issue, we introduce a regularization term to enforce the consistency of gradients by imposing that the gradient's norm be one, as expressed by \(\mathcal{L}_{\textrm{ek}} = \left(\Vert\nabla_{\bm{\theta}} g(\bm{\theta}) \Vert - 1\right)^2\). 
We immediately notice that this is a requirement for being a Signed Distance Field (SDF). 
To integrate the regularization term into the optimization, we take the function \(g\) as the value before the Sigmoid function \(\sigma\), as given by \(f = (\sigma \circ\omega\circ g) (\bm{\theta})\).
The function \( g \) is implemented as a multi-layer perceptron (MLP), \(\omega\) is a scalar multiplication function, and \(\sigma\) is the sigmoid function. 
We notice that although we leverage the SDF, it does not require fitting any manifold distance; the SDF property is ensured by the regularization term. 
Interestingly, the method of imposing the Eikonal constraint is similar to the work of \cite{gropp2020implicit, wang2021neus}; instead of regularizing in three-dimensional space, we regularize the field $g$ in a high-dimensional pose space.

Lastly, we verify in our extensive ablation studies and comparative experiments that our method of learning a high-dimensional pose classifier $f$ can avoid physical collisions with the static environment and the robot itself, making the robotic pose produced by a differentiable pipeline physically reliable.

We summarize our technical contributions as follows:

\begin{enumerate}
    \item We address the issue of efficient differentiable robot rendering with potential collisions involving a static, non-interactive environment and self-collisions. This is achieved by learning a multi-layer perceptron (MLP), denoted as \( f \), to classify the control pose \( \bm{\theta} \) and determine whether it corresponds to a collision. Classification eliminates the need for accurate manifold distance.
   
    \item To address the issue of poor reliability of gradients from the classifier and to integrate \( f \) into a gradient-based optimization pipeline, we introduce the Eikonal regularization term \( \mathcal{L}_{\textrm{ek}} \) to enforce consistency among gradients, effectively transforming the value before the Sigmoid, \( g \), into a signed distance function (SDF). This regularization enhances the reliability of the gradients and leverages the beneficial properties of the SDF without the need for explicit computation of manifold distance.
    
    \item To model the dependency of adjacent robotic joints efficiently, we introduce hierarchical joint encoding that conditions a joint angle on an ancestor's rotation, thereby better capturing the hierarchical dependency of robotic arms.
\end{enumerate}

\section{Related Works}

\subsection{Differentiable Rendering and Simulation}
Differentiable rendering and simulation play a pivotal role in computer vision and graphics.
In differentiable rendering, gradients are propagated from primitives to image, while differentiable simulators compute gradients of physical processes similarly, which facilitates seamless integration with gradient-based optimization techniques.
The integration of neural fields into robotics has attracted considerable attention.  
Researchers have employed diverse representations to bridge the digital and physical realms \cite{yan2016perspective, kato2018neural, henzler2019escaping, liu2019soft, yifan2019differentiable, mescheder2019occupancy, pistilli2020learning, mildenhall2021nerf, park2019deepsdf, kerbl3Dgaussians}.
Innovative approaches, such as Gaussian splatting for the precise reconstruction of robots and their environments, exemplify this trend \cite{lou2024robo, qureshi2024splatsim}.
Moreover, techniques such as neural radiance fields for teleoperation \cite{patil2024radiance} and future scene prediction \cite{lu2025manigaussian} have significantly advanced robotic manipulation capabilities. 
Feature distillation has further empowered robots to perform complex tasks by translating language embeddings into corresponding 3D actions \cite{shen2023distilled, rashid2023language, ze2023gnfactor, qiu2024learning, zheng2024gaussiangrasper}.
Additional advancements include comprehensive physical world modeling, future state prediction  \cite{abou2024physically}, and learning continuous manifolds of valid grasps using signed distance functions (SDF) \cite{khargonkar2023neuralgrasps, weng2023neural}. 
Significant progress has also been made in perceptual scene understanding and grasp planning \cite{breyer2021volumetric, chisari2024centergrasp, dai2023graspnerf}.
One of the primary challenges in developing these simulators is accurately modeling contact interactions—a difficulty that arises from the inherently binary, discontinuous nature of contacts, where objects are either in contact or not \cite{newbury2024review}.
According to \cite{newbury2024review}, contact models can be broadly categorized into four classes: complementarity problems, compliant models, position-based models, and material point method (MPM) frameworks.
Complementarity problems emphasize maintaining strict zero-penetration contact between objects \cite{geilinger2020add, heiden2021neuralsim, howell2022dojo, degrave2019differentiable, Qiao2021Efficient, werling2021fast, gartner2022differentiable, du2021diffpd, li2022diffcloth, le2021differentiable, zhong2021extending, ding2021dynamic, wang2021sim2sim, petrik2022learning, wang2022recurrent, wang2023real2sim2real}. Compliant models relax the complementarity problem's constraint by replacing the sharp, step-like contact functions with smooth approximations that have large but finite gradients\cite{geilinger2020add, du2021diffpd, stuyck2023diffxpbd, gong2022fine, le2023differentiable, turpin2022grasp, turpin2023fast, murthy2020gradsim, heiden2021neuralsim, heiden2019interactive}. The position-based method directly constrains object positions to prevent penetration \cite{freeman2021brax, macklin2022warp}. Meanwhile, the MPM handles contact interactions between both rigid- and soft-body objects\cite{hu2019chainqueen, spielberg2023advanced}.
Despite these impressive advances, the effectiveness of differentiable systems for robotic manipulation remains under active validation, marking a promising yet nascent stage of development.

\subsection{Signed Distance Fields in Robotics}
In the realm of robotics and computer graphics, the construction of SDF has been an important technique for various applications, particularly in motion planning and collision avoidance. Some researchers pioneered a method to construct SDF on grid maps \cite{felzenszwalb2012distance}. Afterwards, some others introduced incremental SDF construction techniques \cite{oleynikova2017voxblox, han2019fiesta}, significantly enhancing the adaptability of SDF in dynamic environments. A novel implicit SDF method is proposed to leverage swept volumes for construction \cite{zhang2023continuous}. This method allows for efficient evaluation of the SDF at specific obstacle points with known coordinates, simplifying the process and reducing computational overhead. Another work \cite{yang2024robotsdf} made a significant stride with the introduction of RobotSDF, an implicit morphology modeling technique. Furthermore, the application of SDF in obstacle avoidance has been extensively explored by \cite{xu2018obstacle, wang2023safe, liu2023collision}.

\section{Method}

\begin{figure*}[h]
    \centering
    \begin{minipage}{0.66\textwidth}
        \centering
        \includegraphics[width=\linewidth]{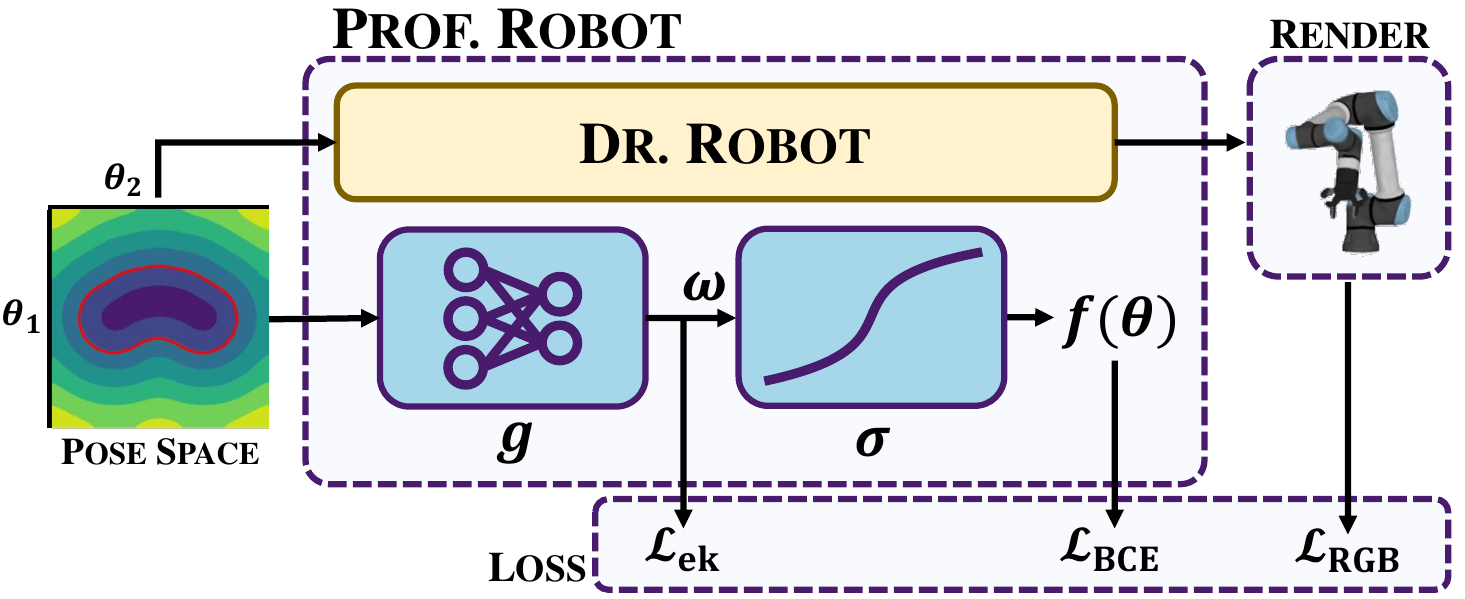}
        \captionof{figure}{\textbf{Our Pipeline.} We predict whether a collision has occurred based on pose input, using Eikonal loss (\(\mathcal{L}_{\textrm{ek}}\)) and binary cross-entropy loss (\(\mathcal{L}_{\textrm{BCE}}\)) to ensure this. It can then be seamlessly integrated with Dr.~Robot~\cite{liu2024differentiable} to create a fully differentiable pipeline for optimization.}
        \label{fig:method}
    \end{minipage}
    \hfill
    \begin{minipage}{0.26\textwidth}
        \centering
        \includegraphics[width=\linewidth]{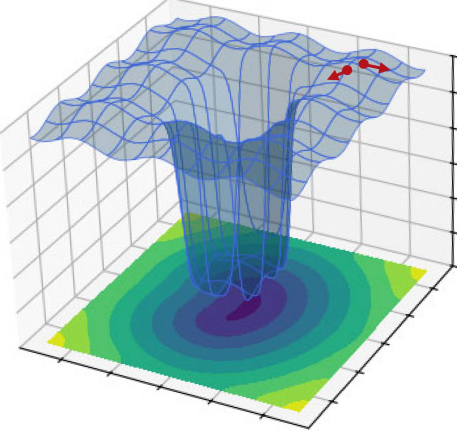}
        \captionof{figure}{\textbf{An illustration of the potential fluctuation of the classification loss landscape.}}
        \label{fig:two-dimensional-f}
    \end{minipage}
\end{figure*}


In this section, we will detail our method for achieving collision avoidance in robotic operations in combination with Gaussian splatting-based differentiable robotic rendering~\cite{liu2024differentiable}. The types of collisions we consider in this paper include static collisions, which occur when the robot collides with a static and non-interactive environment, as well as self-collisions, which involve collisions with the robot itself. Section~\ref{sec:collision_detection_by_pose_classification} will model the problem of collision detection as a robotic pose classification problem. In Section~\ref{sec:eikonal_regularizer_for_consistent_gradients}, we will address some challenges that arose in formulating collision detection and propose the introduction of a signed distance field (SDF) as an underlying field for the classifier to tackle the gradient inconsistency problem during optimization. Section~\ref{sec:hierarchical_robotic_pose_encoding} will explain how we hierarchically encode the robotic pose while respecting the robotic kinematic dependencies. Finally, we will present our loss functions and training schemes in Section~\ref{sec:training_data_and_objectives}.

\subsection{Collision Detection by Pose Classification}
\label{sec:collision_detection_by_pose_classification}
To make the model aware of physical collisions during optimization, we need a collision detector to determine whether the robotic pose has caused intersections with other environmental objects or with the robot itself. The collision detector is essentially a function \( f: \bm{\theta} \mapsto c \) with \( c \in [0, 1] \) that classifies each pose \( \bm{\theta} \) into two categories: either collided (\( c > 0.5 \)) or not (\( c \leq 0.5 \)). 

The robots we consider in this paper have \( K \) joints, and each joint has exactly one degree of freedom for rotation; thus, the pose parameter is within the space \( \bm{\theta} \in \textrm{SO}(2)^K \). The classifier network \( f \) will partition the space into many regions, and ideally, the regions corresponding to valid poses (non-collided poses) will form a manifold as given by

\begin{equation}
    \mathcal{V} = \left\{ \bm{\theta} \in \textrm{SO}(2)^K \mid f(\bm{\theta}) \leq 0.5 \right\}.
\end{equation}


To ensure that the output value falls within the range \([0, 1]\), we follow convention and apply a Sigmoid function \(\sigma\) to the last output, meaning that the function \( f \) is effectively defined by \( f = \sigma \circ g \), where \( g \) is another scalar-valued function that we will detail later in Section~\ref{sec:eikonal_regularizer_for_consistent_gradients}.

To effectively learn the classifier \( f \), we follow tradition and use binary cross-entropy loss to supervise \( f \), as expressed by
\begin{equation}
\mathcal{L}_{\textrm{BCE}} = \textrm{BCE}(f(\bm{\theta}), c^{\textrm{gt}}),
\end{equation}
where \( c^{\textrm{gt}} \in \{0, 1\} \) is the ground truth label indicating whether it is valid (\( c^{\textrm{gt}}=1 \)) or not (\( c^{\textrm{gt}}=0 \)).

\subsection{Eikonal Regularizer for Consistent Gradients}
\label{sec:eikonal_regularizer_for_consistent_gradients}


However, directly applying \( f \) to gradient optimization is problematic. 
The function is nearly binary, and the gradients near the values of \( 0 \) and \( 1 \) are almost zero, leading to noisy and unreliable supervision signals.

We visualize a possible function \( f \) with a two-dimensional input space in Figure~\ref{fig:two-dimensional-f}.
We notice that, since the values at both far sides fluctuate, the gradients also fluctuate, leading to inconsistent gradients. When we aim to optimize a pose from the exterior (collided region) toward the interior (collision-free region), this may not be the case because the gradient directions are inconsistent (see the two red points) due to fluctuations in the loss landscape. Small fluctuations do not significantly influence the loss value, but they can lead to noisy and inconsistent gradients, which are harmful to optimization.

To address the issue, we need to consider the function \( g \) before the Sigmoid function \( \sigma \), as \( \sigma \) will restrict the gradient and render it too small for updating poses. Specifically, for two consecutive gradient updates with poses \( \bm{\theta}_{t} \) and \( \bm{\theta}_{t+1} \), the update at \( t+1 \) is likely to offset the previous update at \( t \), as expressed by 
\begin{equation}
    \left\langle \frac{\partial g}{\partial \bm{\theta}_t}, \frac{\partial g}{\partial \bm{\theta}_{t+1}} \right\rangle \le 0,
\end{equation}
especially in regions with high confidence levels close to \( 0 \) or \( 1 \). This is visualized in Figure~\ref{fig:two-dimensional-f}, where two red dots are plotted; their gradient directions may not be consistent (or correlated) as they point in opposite directions.

We also note that gradient updates are typically small. Assuming the updates \( \bm{\theta}_{t+1} \) and \( \bm{\theta}_{t} \) are close enough and correlated, we have 
\begin{equation}
    \left\langle \frac{\partial g}{\partial \bm{\theta}_t}, \frac{\partial g}{\partial \bm{\theta}_{t+1}} \right\rangle \approx \left\langle \frac{\partial g}{\partial \bm{\theta}_t}, \frac{\partial g}{\partial \bm{\theta}_{t}} \right\rangle = \left\Vert \frac{\partial g}{\partial \bm{\theta}_t} \right\Vert^2 \ge 0.
\end{equation}
This analysis indicates that to enforce gradient consistency, we only need to manipulate the norm of the gradients. To make the correlation of the gradients as uniform as possible (which is desirable for gradient updates), we enforce the above term to be a positive constant \( a > 0 \). This motivates us to adopt the following regularization term
\begin{equation}
    \mathcal{L}_{\textrm{ek}} = \left( \left\Vert \frac{\partial g}{\partial \bm{\theta}} \right\Vert - a \right)^2
\end{equation}
to ensure gradient consistency between the two adjacent updates. We immediately notice that when \( a = 1 \), it becomes an Eikonal regularizer that is widely employed by the signed distance function~\cite{park2019deepsdf, gropp2020implicit}. Interestingly, this term can be viewed as generalizing the implicit geometric regularization~\cite{gropp2020implicit} to high-dimensional pose space by learning a function \( g \). To simplify implementation, we use \( a = 1 \) throughout the paper.

The function \( g \) is practically implemented as a multi-layer perceptron that receives robotic poses \( \bm{\theta} \) as input and produces a scalar; the final classifier is therefore 
\begin{equation}
    f(\bm{\theta}) = (\sigma \circ \omega \circ g)(\bm{\theta}),
\end{equation}
where \(\omega: x \mapsto sx\) denotes a function that multiplies the input \(x\) by an optimizable positive constant \(s\), which will be further discussed in Section \ref{sec:scaling_factor}. Our current pipeline is illustrated in Figure~\ref{fig:method}.






\subsection{Hierarchical Robotic Pose Encoding}
\label{sec:hierarchical_robotic_pose_encoding}
We represent the robotic pose using rotation angles, each describing the relative rotation between two arms. The pose parameters are defined in local coordinate frames, such that continuous modifications of a single joint result in realistic motion. However, since angles only describe relative rotations, they may not respect the kinematic structure and cannot effectively handle the structural dependency relationships between the arms. We observe that each individual joint needs to be conditioned on the parent rotations and thus must be influenced by them. To incorporate this notion of kinematic dependency into the function \(g\), we implement the function as a hierarchical network that encodes the dependency within the network structure~\cite{aksan2019structured, georgakis2020hierarchical, mihajlovic2021leap, tiwari2022pose}.

This is similar to the method described in \cite{tiwari2022pose}.
Specifically, for a given robotic pose \(\bm{\theta} = [\theta_1; \ldots; \theta_K]\), where \(\theta_k\) (with \(k \in \{1, \ldots, K\}\)) is the rotation angle for joint \(k\), and a function \(\tau(k)\) maps the index of each joint to its parent joint index, we encode each pose using a multi-layer perceptron (MLP) as follows:
\begin{equation}
\begin{aligned}
    g_1 &: \theta_1 \mapsto \bm{\nu}_1 \\
    g_k &: (\theta_k, \bm{\nu}_{\tau(k)}) \mapsto \bm{\nu}_k, \quad k \in \{2, \ldots, K\}
\end{aligned}
\end{equation}
This process takes the joint angle and the encoded feature \(\bm{\nu}_{\tau(k)} \in \mathbb{R}^\ell\) of its parent joint as input and produces an output feature \(\bm{\nu}_k \in \mathbb{R}^\ell\), where \(\ell = 8\) is the feature dimension. We then concatenate the encoded features of every joint to obtain a combined pose embedding \(\bm{\nu} = \left[\bm{\nu}_1; \ldots; \bm{\nu}_K\right]\). This embedding is further transformed by a linear mapping \(\mathbb{R}^{\ell \cdot K} \to \mathbb{R}\) to yield a scalar value that serves as the output of the function \(g\).


\subsection{Training Data and Objectives}
\label{sec:training_data_and_objectives}
The training dataset comprises \( N \) poses and their corresponding decisions, represented by \( \mathcal{D} = \{(\bm{\theta}_i, c^{\textrm{gt}}_i)\}_{i=1}^{N} \). It is constructed as follows: the robotic pose \( \bm{\theta}_i \) is randomly sampled, and a (non-differentiable) collision detector is then applied to produce the decision \( c^{\textrm{gt}}_i \in \{0, 1\} \), indicating whether a collision has occurred (\( c^{\textrm{gt}}_i = 1 \)).

The model is trained using batched samples; we omit the summation-over-samples notation here for simplicity. The objective is simply the weighted combination of \( \mathcal{L}_{\textrm{BCE}} \) and \( \mathcal{L}_{\textrm{ek}} \). With \( \alpha = 0.1 \), it is expressed as 
\begin{equation}
    \mathcal{L} = \mathcal{L}_{\textrm{BCE}} + \alpha \mathcal{L}_{\textrm{ek}}.
\end{equation}
Depending on how it is learned, we may additionally have, for example, \( \mathcal{L}_{\textrm{RGB}} \) for image-space supervision, which we omit here for simplicity.






\section{Experiments and Results}

We conduct ablation studies on Prof.~Robot (section \ref{sec:learning-robotic-signed-distance-fields}), experiments on collision resolution (section \ref{sec:collision-resolution-by-pose-optimization}), integration with Dr.~Robot (section \ref{sec:integrated_with_differentiable}), and sim-to-real deployment (section \ref{sec:sim-to-real-deployment}). 
Comparison with the differentiable simulator, training with alternative regularizers, and manipulation with Dr.~Robot are detailed in the supplementary materials.

\subsection{Experimental Setup}

\noindent\textbf{Data Preparation.} We adopt a data-driven approach, leveraging the Mujoco \cite{todorov2012mujoco} configurations provided by \cite{liu2024differentiable} with some modifications. Specifically, we selected six scenes and randomly sampled poses from the robot's joint limits, used Mujoco (which is non-differentiable) to check for collisions, and employed a resampling technique to balance the ratio of collision to non-collision data at approximately the same amount.

\noindent\textbf{Evaluation Metrics.} We primarily evaluate the network's ability to distinguish between collision and non-collision states, as well as its performance in optimizing the gradients. The former metric (accuracy) assesses the network's ability to detect collisions, while the latter (\(\mathcal{L}_{\textrm{ek}}\)) measures its capacity to avoid collisions through optimization. Furthermore, for a given robot pose denoted as $\bm\theta$, we are able to determine the collision points of the mesh within the simulator and subsequently calculate the collision rate for all possible poses.

\noindent\textbf{Experimental Details.} 
We leverage the joint connection relationships defined in the URDF, where each joint is encoded using a tiny MLP. The encoded features from all joints are then concatenated and passed through an MLP with four hidden layers, each containing 512 units, followed by a final output layer with a single unit. LeakyReLU is used as the activation function for all layers. We introduce a learnable scale factor \(s\), which, after being multiplied by the MLP output, is passed through a Sigmoid function to produce a probability. 
We conduct experiments within a simulation environment, where the network's ability determines the success rate of collision resolution. Finally, we integrate our method into a differentiable robot rendering pipeline and perform control experiments. The experiments are conducted on a single NVIDIA RTX 4090 GPU.

\subsection{Learning Robotic Signed Distance Fields}
\label{sec:learning-robotic-signed-distance-fields}

\subsubsection{Ablation: Hierarchical Encoding}

To assess the impact of hierarchical encoding, we replaced it with a simple linear layer for encoding the input joint data while keeping the rest of the components unchanged. As shown in Table~\ref{tab:comp_bet_hire_a_fla_enc}, after 10,000 iterations, the use of the hierarchical MLP results in higher accuracy. This shows that the hierarchical encoding strategy better captures the kinematic structure and increases performance.

\begin{table}
\small
\caption{\textbf{Ablation study on hierarchical and flattened encoding.}}
    \centering
    \begin{tabular}{c|cc}
        \toprule
                 & Hierarchical       & Flatten \\
        \midrule
        Classification Accuracy & \textbf{97.70\%}   & 97.51\% \\  
        \bottomrule
    \end{tabular}
    \vspace{-0.2cm}
    \label{tab:comp_bet_hire_a_fla_enc}
\end{table}

\subsubsection{Ablation: Eikonal Regularizer}

To investigate the effect of the Eikonal regularizer, we trained an MLP under two conditions: one without the Eikonal regularizer and one with it, and compared both the classification accuracy and the numerical value of the Eikonal loss. We then tested whether the models could resolve collisions (as described in Section~\ref{sec:collision-resolution-by-pose-optimization}). The models were trained for 10,000 iterations. During pose optimization, for the model trained without the Eikonal regularizer, we stopped the iteration if the collision probability dropped below 0.5. 
For the model trained with the Eikonal regularizer, we halted the iteration if the SDF value, which is negative in non-collision regions, fell below the threshold \(l_{\text{thres}} = -0.1\).

The results were compared, as shown in Table~\ref{tab:comp_bet_w_a_wo_eki_loss}.
Without the Eikonal regularizer, the optimizer struggled to generate a collision-free pose from a collision state, with iterations exceeding 1,000, and the propagated gradients were too small to facilitate further optimization.

\begin{table}
\small
\caption{\textbf{Ablation study on the Eikonal loss.}}
    \centering
    \begin{tabular}{c|cc}
        \toprule
                        & w. Eikonal loss      & w/o. Eikonal loss    \\
        \midrule
        \makecell{{\footnotesize Classification}\\ {\footnotesize Accuracy}}        & 97.70\%              & \textbf{98.39\%}     \\
        Iteration Nums $\downarrow$  & \textbf{119.99}      & 1000                 \\
        Iteration Time $\downarrow$  & \textbf{0.97 s}      & 10.64 s              \\
        Move Angle $\downarrow$      & \textbf{0.60}        & 4.49                 \\
        Collision Point $\downarrow$ & \textbf{0.67}        & 11.52                \\
        Collision Rate $\downarrow$  & \textbf{17\%}        & 99\%                 \\
        Gradient Norm   & $\bm{8.53\times 10^{-1}}$ & $1.89\times 10^{-6}$ \\
        \bottomrule
    \end{tabular}
    \vspace{-0.2cm}
    \label{tab:comp_bet_w_a_wo_eki_loss}
\end{table}

\subsubsection{Ablation: Scaling Factor}
\label{sec:scaling_factor}

The cross-entropy loss and Eikonal loss may require gradients of different magnitudes, potentially interfering with each other and hindering the model's ability to learn useful information. To address this, a scaling factor \(s\) for the SDF is introduced to allow for large gradients in classification and limited gradients in the SDF. We removed the scaling factor \(s\) (setting \(s = 1\)) and directly passed the SDF value to the Sigmoid function to produce \(f\). 

The numerical results, as shown in Table~\ref{tab:comp_bet_w_a_wo_s}, demonstrate that the scaling factor effectively achieves good classification results while preserving the uniform gradient property.

\begin{table}
\small
\caption{\textbf{Ablation study on the scaling factor $s$.}}
    \centering
    \begin{tabular}{c|cc}
        \toprule
               & Accuracy         & Eikonal Loss $\downarrow$    \\
        \midrule
        w. $s$   & \textbf{97.70\%} & \textbf{0.0850} \\
        w/o. $s$ & 85.37\%          & 0.2568          \\
         \bottomrule
    \end{tabular}
    \vspace{-0.2cm}
    \label{tab:comp_bet_w_a_wo_s}
\end{table}

\subsubsection{Comparison Between NDF and SDF}

In contrast to \cite{tiwari2022pose}, which models collision-free poses as a high-dimensional submanifold using a Neural Distance Field (NDF), we represent collision-free poses as a set of discontinuous regions in high-dimensional space that can be modeled using an SDF. To explore the differences between these two approaches, we also conducted experiments with an NDF. Learning an NDF requires accurate manifold distances; unlike \cite{tiwari2022pose}, our method does not require this because it can be implicitly learned through \(\mathcal{L}_{\textrm{ek}}\). 
To provide distance labels for \cite{tiwari2022pose}, we followed their convention and employed Faiss to compute the L2 distance of each pose as an approximate ground truth distance.

After 10,000 iterations of optimization, the results are presented in Table~\ref{tab:compare_between_learning_of_sdf_and_udf}. As shown in the table, our modeling approach outperforms the NDF. By learning a region (rather than a submanifold) and implicitly learning through \(\mathcal{L}_{\textrm{ek}}\), our method achieves better results.

\begin{table}
\small
\caption{\textbf{Comparison between SDF and NDF.}}
    \centering
    \begin{tabular}{c|cc}
        \toprule
             & Accuracy          & Eikonal Loss $\downarrow$\\
        \midrule
        SDF  & \textbf{97.70\%}  & \textbf{0.0850} \\
        NDF  & 54.33\%           & 0.1916 \\
         \bottomrule
    \end{tabular}
    \vspace{-0.2cm}
    \label{tab:compare_between_learning_of_sdf_and_udf}
\end{table}

\subsection{Collision Resolution by Pose Optimization}
\label{sec:collision-resolution-by-pose-optimization}

\begin{figure}[h]
    \centering
    \includegraphics[width=0.8\linewidth]{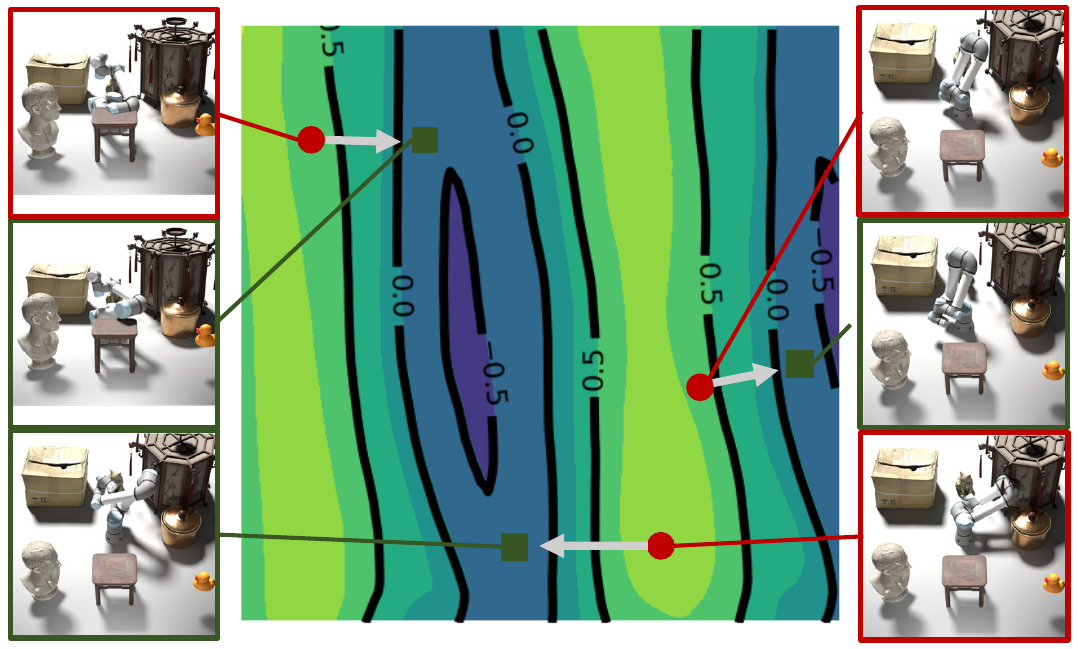}
    \captionof{figure}{\textbf{An illustration of how collision resolution optimizes a collided robot towards a collision-free state.} The figure shown visualizes the SDF, where red points (representing collisions) are optimized to green points (representing non-collisions) along the gradient update trajectories.}
    \label{fig:gradient_optimization_for_collision_free_poses}
\end{figure}

Given a collision pose, our neural network can optimize the collision pose to the nearest non-colliding pose, as shown in Figure~\ref{fig:gradient_optimization_for_collision_free_poses}. After training the network, we sample 1,000 collision poses and use our $g$ to predict their SDF values. We then apply gradient optimization, fixing the trained model's weights, and minimize the predicted SDF until it falls below a threshold value $l_\textrm{thres}$. 

Specifically, we use the Adam optimizer~\cite{kingma2014adam} with a learning rate of 0.01, keeping all other parameters unchanged. The results are presented in Table~\ref{tab:optimization_of_collision_free_pose_by_sdf}. As the SDF values decrease, both the likelihood and severity of collisions reduce, demonstrating that our method effectively predicts and avoids collisions.

\begin{table}
\small
\caption{\textbf{Effect of different threshold ($l_\textrm{thres}$).} 
\(\overline{\text{Iter.}}\) represents the average number of iterations, while \(\Delta\theta\) denotes the change in angle.
The last two columns display the collision points in the scenario prior to optimization and after optimization, respectively. }
    \centering
    \begin{tabular}{c|ccc|cc}
        \toprule
        $l_\textrm{thres}$ & \(\overline{\text{Iter.}}\) $\downarrow$ & Time (s) $\downarrow$        & \(\Delta\theta\) $\downarrow$          & Before & After $\downarrow$         \\
        \midrule
        -0.000             & 19.017          & 0.155          & 0.162          & 5.557 & 0.399          \\
        -0.001             & 19.392          & 0.156          & 0.165          & 5.633 & 0.375          \\
        -0.005             & \textbf{18.874} & \textbf{0.153} & \textbf{0.159} & 5.508 & 0.312          \\
        -0.010             & 20.124          & 0.154          & 0.173          & 5.617 & 0.187          \\
        -0.020             & 20.090          & 0.162          & 0.173          & 5.471 & 0.080          \\
        -0.030             & 20.236          & 0.160          & 0.172          & 5.343 & 0.019          \\
        -0.040             & 21.256          & 0.171          & 0.181          & 5.626 & 0.016          \\
        -0.050             & 22.627          & 0.174          & 0.190          & 5.757 & 0.006          \\
        -0.100             & 24.450          & 0.192          & 0.203          & 5.197 & \textbf{0.001} \\
        \bottomrule
    \end{tabular}
    \vspace{-0.2cm}
    \label{tab:optimization_of_collision_free_pose_by_sdf}
\end{table}

To assess the robustness of our model, we perform optimization with varying learning rates, anticipating consistent performance across these rates. To examine the effect of the learning rate, we replicate the experimental setup as described previously, setting \( l_\textrm{thres} = -0.1 \). We then compare the number of collision points and analyze the overall trends. The results demonstrate that the model performs comparably across different learning rates, further validating the robustness of our approach. A summary of these results is presented in Table~\ref{tab:impact_of_different_learning_rate}. 

\begin{table}
\small
\caption{\textbf{Impact of different learning rate.} The content of the last two columns is identical to that of Table~\ref{tab:optimization_of_collision_free_pose_by_sdf}.}
    \centering
    \begin{tabular}{c|ccc|cc}
        \toprule
        LR    & \(\overline{\text{Iter.}}\) $\downarrow$        & Time(s) $\downarrow$        & \(\Delta\theta\) $\downarrow$          & Before & After $\downarrow$         \\
        \midrule
        0.001 & 241.363          & 1.841          & \textbf{0.203} & 5.460 & 0.003          \\
        0.010 &  24.450          & 0.192          & \textbf{0.203} & 5.197 & 0.001          \\
        0.020 &  12.826          & 0.104          & 0.215          & 5.495 & \textbf{0.000} \\
        0.050 &   5.634          & 0.049          & 0.236          & 5.622 & \textbf{0.000} \\
        0.100 &   3.026          & 0.028          & 0.256          & 5.569 & \textbf{0.000} \\
        0.200 &   1.785          & 0.018          & 0.315          & 5.635 & 0.001          \\
        0.500 &   1.127          & \textbf{0.013} & 0.527          & 5.326 & \textbf{0.000} \\
        1.000 &   \textbf{1.104} & \textbf{0.013} & 1.040          & 5.553 & \textbf{0.000} \\
        \bottomrule
    \end{tabular}
    \label{tab:impact_of_different_learning_rate}
    \vspace{-0.2cm}
\end{table}

\subsection{Integration with Dr.~Robot}
\label{sec:integrated_with_differentiable}

Differentiable robot rendering (Dr.~Robot) enables us to obtain a differentiable pose from the rendered view. From this pose, our method can infer prior knowledge about potential collisions with the external environment or self-intersections. This prior knowledge can be seamlessly integrated into the differentiable rendering process.

\subsubsection{Inverse Action from Collision Image}
\label{sec:inverse_action_from_collision_image}

We leverage Dr.~Robot to optimize the robot's pose from images. However, we observe that the optimized poses are not always reasonable; in some cases, they result in collisions, limiting the practical application of Dr.~Robot in real-world scenarios. To address this, we combine our SDF $g$ with Dr.~Robot, enabling us to optimize reasonable poses that minimize collisions while maintaining image consistency.

Striving for an efficient evaluation, we introduce immovable rigid blocks next to the robot. This way, if the robot's pose is unreasonable, it will collide with these blocks or self-collision. In each scene, we set up four cameras from different viewpoints and capture rendered images, instance segmentations, and depth maps from the simulation engine. This setting closely aligns with the real-world scenarios using monocular depth estimation models, such as Depth Pro\cite{depth-pro}. To optimize the robot's pose, we use the following loss functions:
\begin{equation}
\mathcal{L}_\textrm{RGB} = \frac{1}{N}\sum_{\textrm{seg}\left[i, j\right]=1}\|\bm r-\bm r_\textrm{gt}\|    
\end{equation}
\begin{equation}
\mathcal{L}_\textrm{D} = \frac{1}{N}\sum_{\textrm{seg}\left[i, j\right]=1}\|\bm d - \bm d_\textrm{gt}\|    
\end{equation}
Where $\textrm{seg}\left[i, j\right]$ is the robot's mask, with 0 indicating the robot is not present at that position in the image and 1 indicating it is present. Here, $\bm r$ represents the rendered image from Dr.~Robot, and $\bm r_\textrm{gt}$ represents the rendered image from the simulation engine; $\bm d$ represents the rendered depth from Dr.~Robot, and $\bm d_\textrm{gt}$ represents the rendered depth from the simulation engine. The baseline loss function is the sum of the image and depth losses:
\begin{equation}
\mathcal{L} = \mathcal{L}_\textrm{RGB} + \mathcal{L}_\textrm{D}    
\end{equation}
To address collisions, we use our trained SDF model. The data burden is manageable as we can obtain sufficient data from the simulation engine for training. During the optimization process, we introduce an additional term to the loss function:
\begin{equation}
\mathcal{L}_\textrm{SDF} = \textrm{ReLU}\left(g\left(\bm\theta\right)-l_\textrm{thres}\right)
\label{eq:l_sdf}
\end{equation}

Thus, our loss function is
\begin{equation}
\mathcal{L} = \mathcal{L}_\textrm{RGB} + \mathcal{L}_\textrm{D} + \beta \cdot \mathcal{L}_\textrm{SDF}    
\end{equation}
where $\beta$ is a weight scalar. The optimization stops when $g(\bm\theta) \le l_\textrm{thres}$, with $l_\textrm{thres}=-0.1$. We sampled 100 instances for testing, and the results are shown in Table~\ref{tab:collision_rate_of_the_pose_inverse_from_images}. The quantitative results demonstrate that, after optimization, Dr.~Robot effectively avoids collisions and achieves the closest possible pose without collision.

\begin{table}
\small
\caption{\textbf{The number of collision points (averaged) and collision rate of the pose inverse from images.} We control the collision point from simulator and generate images for Dr.~Robot optimization.}
\vspace{-0.2cm}
    \centering
    \begin{tabular}{c|cc|cc}
        \toprule
        Original collision & \multicolumn{2}{c|}{Collision Point $\downarrow$} & \multicolumn{2}{c}{Collision Rate $\downarrow$}  \\
        point in data   & Dr.~Robot & Ours           & Dr.~Robot & Ours          \\
        \midrule
        10.96           & 9.17      & \textbf{0.83}  & 78\%      & \textbf{9\% } \\
         2.00           & 2.33      & \textbf{0.04}  & 86\%      & \textbf{2\%}  \\
         1.00           & 1.18      & \textbf{0.02}  & 78\%      & \textbf{2\%}  \\
         0.00           & 0.05      & \textbf{ 0.00} &  2\%      & \textbf{0\%}  \\
        \bottomrule
    \end{tabular}
    \vspace{-0.2cm}
    \label{tab:collision_rate_of_the_pose_inverse_from_images}
\end{table}

\subsubsection{Inverse Trajectory from Optimization}
\label{sec:inverse_trajectory_from_opt}


Furthermore, we aim to explore the feasibility of using Dr.~Robot for trajectory generation. Given the starting posture \(\bm{\theta}_\textrm{start}\) and the end posture \(\bm{\theta}_\textrm{end}\), we employ gradient optimization with Dr.~Robot to generate the intermediate trajectory for robot control.

As mentioned in Section~\ref{sec:inverse_action_from_collision_image}, we randomly sample two collision-free poses and obtain their respective rendering results. Dr.~Robot uses \(\bm{\theta}_\textrm{start}\) as the initial configuration and utilizes the image, mask, depth, and pose information of \(\bm{\theta}_\textrm{end}\) from the simulator as supervisory signals. In addition, we incorporate our SDF model and add \(\mathcal{L}_\textrm{SDF}\) (Equation \ref{eq:l_sdf}) to the loss function.

However, during the optimization process, the value of the SDF fluctuates around \(l_{\textrm{thres}}\), which causes some data points to violate the SDF constraints. To address this, we perform post-processing on the intermediate trajectory:
\begin{equation}
\mathcal{L} = \frac{1}{N}\sum_{i=1}^{N}\mathcal{L}_\textrm{SDF}\left(\hat{\bm\theta_i}\right) + \frac{1}{N-1}\sum_{i=1}^{N-1}\|\hat{\bm{\theta}}_{i+1}-\hat{\bm{\theta}_{i}}\|    
\end{equation}
where $\hat{\bm\theta_i}, i=1 \ldots N$ is the middle trajectory generate from optimization. Here, we combine the SDF loss and the total variation loss. The results, as shown in Table~\ref{tab:optimize_trajectory_by_drrobot_and_sdf}, demonstrate that our approach effectively generates intermediate trajectories suitable for robot control.

\begin{table}
\small
\caption{\textbf{Optimize Trajectory by Dr.~Robot and SDF.}}
    \centering
    \begin{tabular}{c|cc}
        \toprule
                    & Collision Point $\downarrow$ & Collision Rate $\downarrow$   \\
        \midrule
        Dr.~Robot   & 4.07           & 10.64\%         \\
        Ours (-0.00) & 3.50           & 10.00\%         \\
        Ours (-0.01) & \textbf{0.00}  & \textbf{0.00\%} \\
        Ours (-0.05) & \textbf{0.00}  & \textbf{0.00\%} \\
        Ours (-0.10) & \textbf{0.00}  & \textbf{0.00\%} \\
        \bottomrule
    \end{tabular}
    \vspace{-0.2cm}
    \label{tab:optimize_trajectory_by_drrobot_and_sdf}
\end{table}

\subsubsection{Inverse Action from VLM}

We have observed that directly extracting data from the Vision-Language Model (VLM) using Dr.~Robot can lead to unconstrained optimization, potentially resulting in implausible poses. Conversely, employing a simulator to refine these poses is problematic due to the absence of gradient information, which hinders seamless integration with Dr.~Robot. Our novel approach successfully circumvents the optimization of collision-prone poses, thereby maintaining the plausibility of the outcomes.

In alignment with Dr.~Robot's methodology, we utilized CLIP to extract action data from the VLM. Our experiments use \textit{shadow hand} and leveraged the \textit{openai/clip-vit-base-patch32} model from HuggingFace. We aimed to maximize the dot product between the language and image embeddings generated by their respective processing towers. We computed the optimized collision points and associated collision probabilities. To substantiate that our method does not compromise the efficacy of action distillation from the VLM, we conducted a comparative analysis of the optimized dot products. The findings, presented in Table~\ref{tab:collision_result_from_dr_robot_and_sdf_by_clip}, demonstrate that our technique effectively precludes the optimization of implausible poses without impeding the extraction of actions from the VLM.

\begin{table}
\small
\caption{\textbf{The metrics of collision and optimization while optimizing robotic pose by CLIP.} We indicate the collision point(Point) and collision rate(Rate) for average, and report the mean and std of dot product metrics.}
    \centering
    \begin{tabular}{c|cc|cc}
        \toprule
                    & Point $\downarrow$ &  Rate $\downarrow$ & mean           & std $\downarrow$      \\
        \midrule
        Dr.~Robot   & 1.30           & 42.50\%          & 33.86          & 1.53          \\
        Ours (-0.00) & 0.40           & 30.00\%          & \textbf{34.42} & 1.58          \\
        Ours (-0.01) & 0.80           & 40.00\%          & 34.13          & \textbf{1.20} \\
        Ours (-0.05) & 0.30           & 20.00\%          & 34.22          & 1.60          \\
        Ours (-0.10) & \textbf{0.10}  & \textbf{10.00\%} & 34.02          & 1.23          \\
        \bottomrule
    \end{tabular}
    \vspace{-0.2cm}
    \label{tab:collision_result_from_dr_robot_and_sdf_by_clip}
\end{table}

\subsection{Sim-to-Real Deployment}
\label{sec:sim-to-real-deployment}

\begin{figure}
    \centering
    \includegraphics[width=\linewidth]{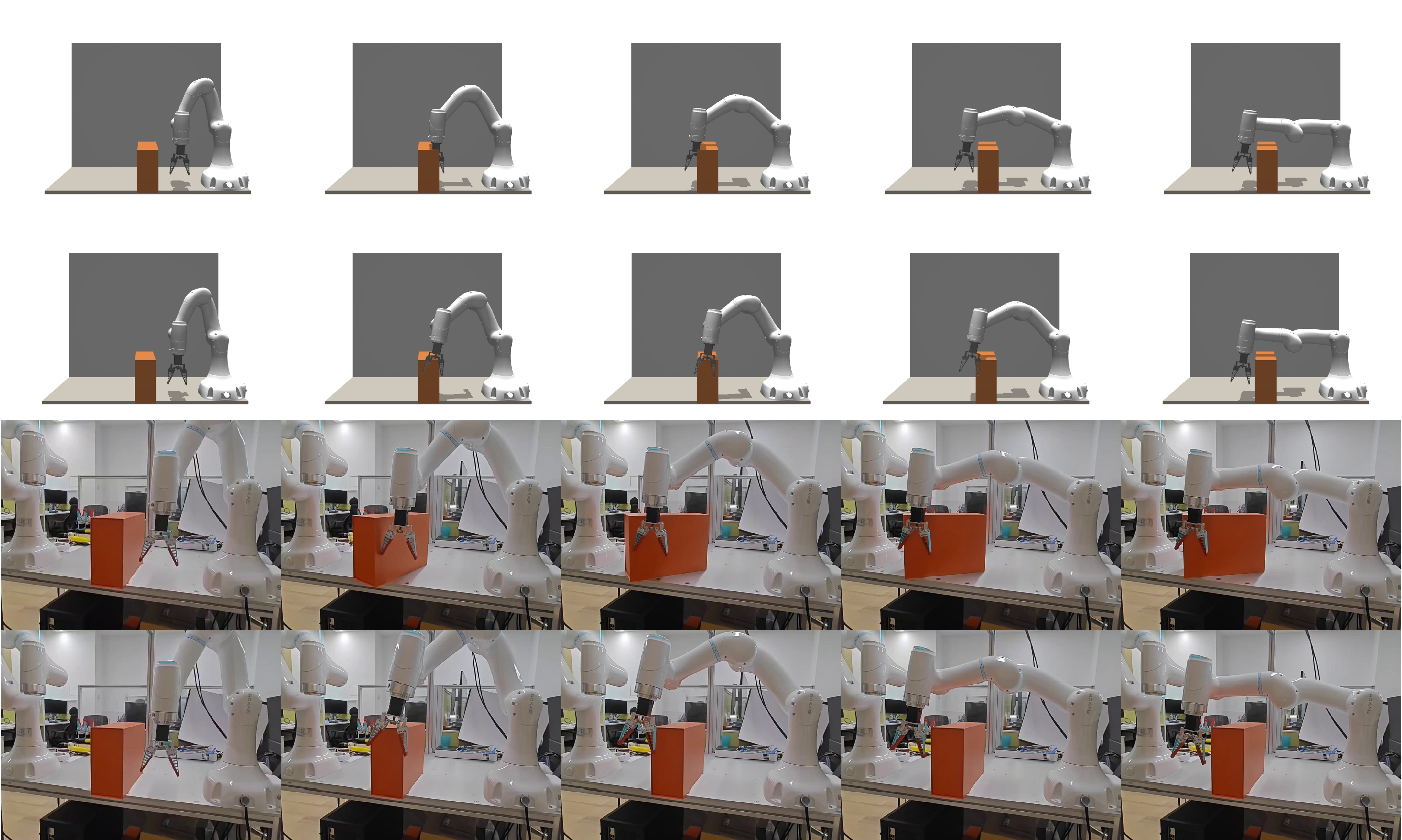}
    \caption{\textbf{Sim2Real Experiment: Obstacle Avoidance with Prof.~Robot.} The first row illustrates that, in the simulator, the robot arm collides with the red box due to its inability to perceive obstacles. The second row shows that after integrating Prof.~Robot, the robot successfully avoids the red obstacle. The red box represents an obstacle, and without Prof.~Robot, the robot fails to circumvent it. The final row demonstrates that, with Prof.~Robot, the robot smoothly navigates around the obstacle, confirming the effectiveness of our method for real-world applications.}
    \label{fig:sim2real}
    \vspace{-0.45cm}
\end{figure}

To validate the performance of our method, we conducted several sim-to-real experiments.
In an illustrative scenario, a robot navigates around obstacles -- including a wall and a box -- while moving from an initial pose to a target pose.
We interpolate 100 poses between the initial and target positions using SO(2) interpolation and optimize the trajectory with our Prof.~Robot algorithm.
The results, shown in Figure \ref{fig:sim2real}, demonstrate that our approach effectively guides the robot arm around obstacles while preventing collisions in simulation and reality.



\section{Limitations and Future Work}

Our method tackles the problem of cross-environment generalization. 
To address this, a new mechanism could be proposed that enhances Prof.~Robot's scene perception by modeling dynamic objects.
The dynamic environment could be represented by supplementary joints \(\delta\), which constrain the robotic system's joint space.
By training with a collision-aware discriminator \(p(\cdot|\delta)\), our joint learning framework could practical effectiveness despite environmental perturbations from ideal conditions.
In practical scenarios, this is achieved by estimating the poses of dynamic objects.

{
    \small
    \bibliographystyle{ieeenat_fullname}
    \bibliography{main}
}

\clearpage
\setcounter{page}{1}
\maketitlesupplementary

\section{Network Architecture}
\label{sec:network-architecture}

\begin{figure}[h]
    \centering
    \includegraphics[width=0.8\linewidth]{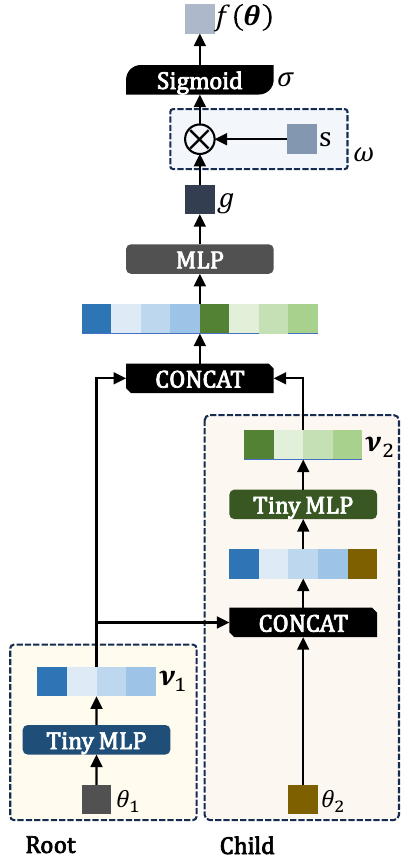}
    \captionof{figure}{Architecture}
    \label{fig:architecture}
\end{figure}

The architecture of our network, illustrated in Figure \ref{fig:architecture}, presents an elegantly straightforward design that comprises two distinct inputs and a singular dependencies.

\section{Collision Resolution}

\begin{figure*}
    \centering
    \includegraphics[width=0.8\textwidth]{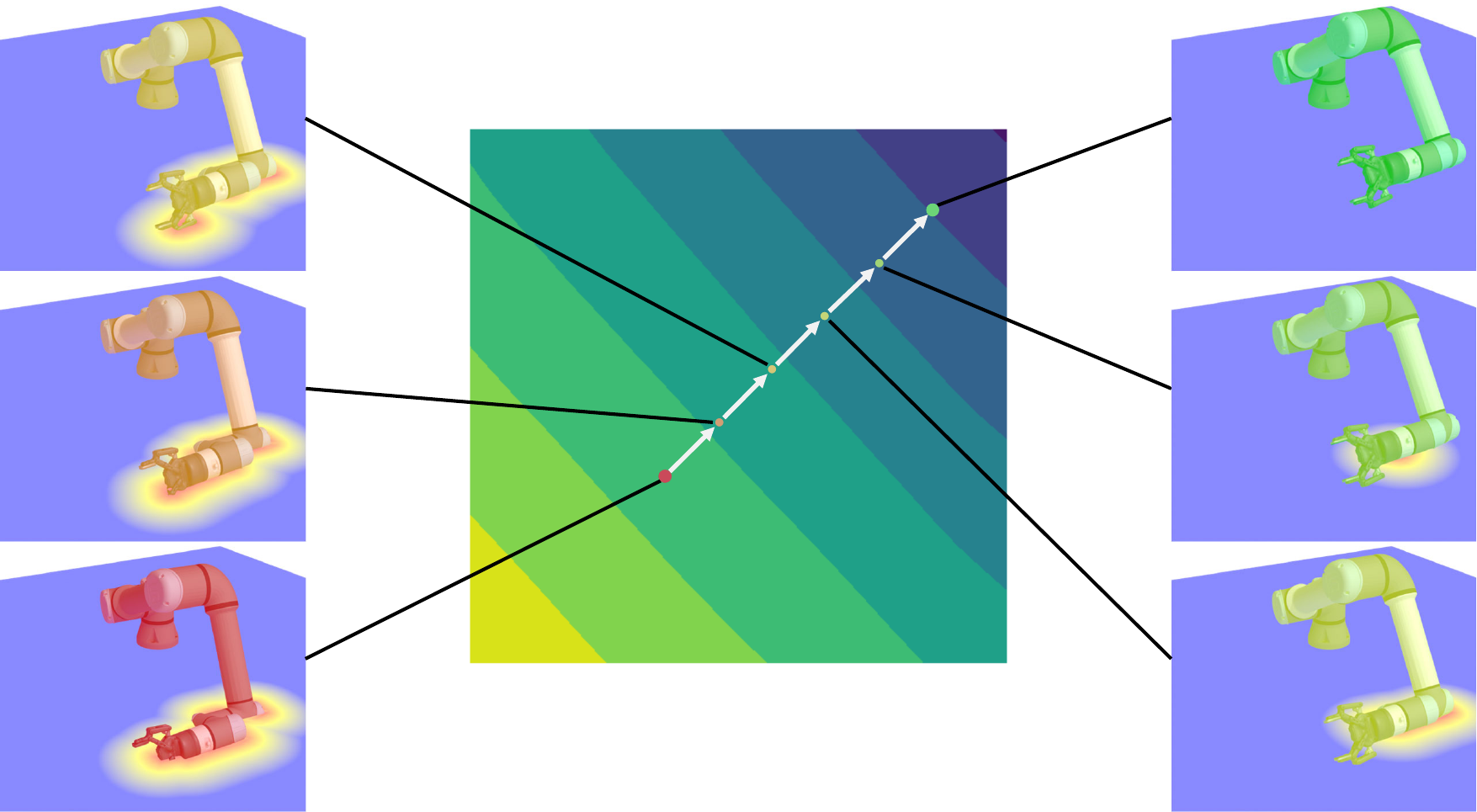}
    \captionof{figure}{\textbf{The Visualization of SDF Optimization}. We adjust the angles of only two of the robot's joints while keeping the remaining joint angles fixed. Using these two joint angles as coordinate axes, we calculate the SDF values corresponding to various joint configurations. Subsequently, we apply the optimization method detailed in Section \ref{sec:collision-resolution-by-pose-optimization}, visualizing the joint data points alongside their corresponding states of the robot in a plot. In this visualization, the robot transitions from {\color{red} \textbf{red}} collision points to {\color{green} \textbf{green}} non-collision points. For improved clarity, the regions where the robot comes into contact with the plane are depicted in {\color{red} \textbf{red}}, the collision-free areas in {\color{blue} \textbf{blue}}, and the transitional zones in {\color{yellow} \textbf{yellow}}.}
    \label{fig:visualization-of-sdf-optimization}
\end{figure*}

\begin{figure}[h]
    \centering
    \includegraphics[width=0.9\linewidth]{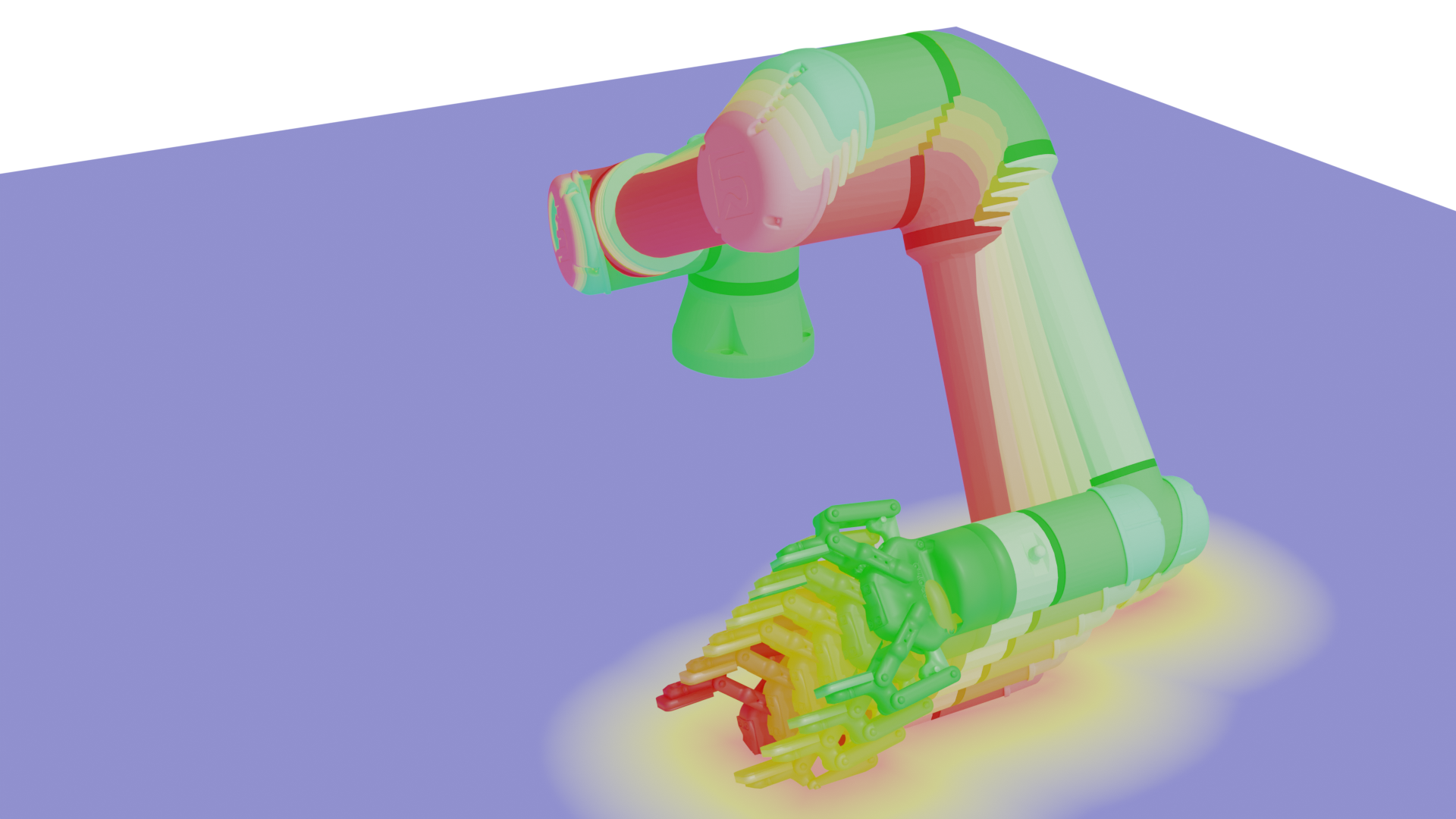}
    \captionof{figure}{\textbf{The Stack of SDF Optimization}. To effectively illustrate the robot's gradual transition from a collision state to a non-collision state, we stack the intermediate results from the optimization process shown in Figure \ref{fig:visualization-of-sdf-optimization} for visualization purposes.}
    \label{fig:stack-of-the-sdf-optimization}
\end{figure}

We illustrate the process of optimizing the robot's joint angles for collision resolution in Figure \ref{fig:visualization-of-sdf-optimization}. This transformation is further clarified by layering intermediate optimization results in Figure \ref{fig:stack-of-the-sdf-optimization}, thus offering a clearer depiction of how the robot gradually evolves from an initial state of collision to a final, non-collision posture.

\section{Collision-free Trajectory}

As outlined in Section \ref{sec:inverse_trajectory_from_opt}, given the initial posture \(\bm\theta_\textrm{start}\) and the target posture \(\bm\theta_\textrm{end}\), we can delineate the trajectory for the transition from \(\bm\theta_\textrm{start}\) to \(\bm\theta_\textrm{end}\). We illustrate this process using three distinct methodologies.

\subsection{Trajectory of Interpolation}

Interpolation within the \(\textrm{SO}(2)\) plane offers a naive solution. Assuming the interpolation function in \(\textrm{SO}(2)\) is represented as \(\mathcal{I}\), we can derive \(N\) points along the intermediate trajectory, \(\hat{\bm\theta_1}, \ldots, \hat{\bm\theta_N}\), defined by
\begin{equation}
    \hat{\bm\theta_1}, \ldots, \hat{\bm\theta_N} = \mathcal{I}\left( \bm\theta_\textrm{start}, \bm\theta_\textrm{end} \right )
\end{equation}

\begin{figure*}
    \centering
    \includegraphics[width=\textwidth]{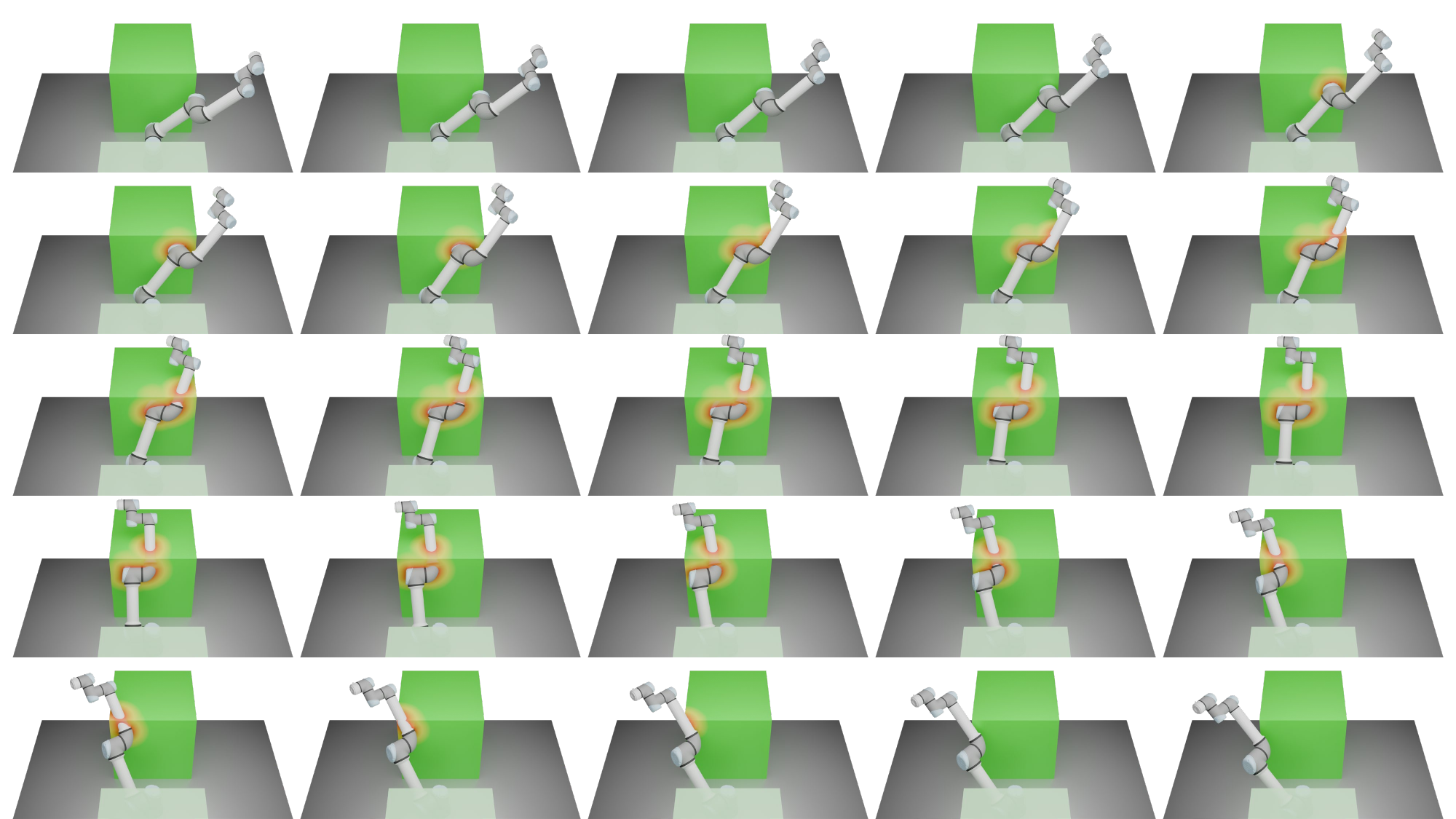}
    \captionof{figure}{\textbf{Visualization of the interpolation trajectory.} The reading sequence progresses from left to right and top to bottom, adhering to this pattern consistently. When the robotic arm makes contact with the {\color{green} \textbf{green}} block, the collision area will be highlighted in {\color{red} \textbf{red}}, with a {\color{yellow} \textbf{yellow}} gradient transitioning in between, a pattern that is replicated in other instances as well.}
    \label{fig:visualization-of-interpolation-trajectory}
\end{figure*}

\begin{figure*}
    \centering
    \includegraphics[width=\textwidth]{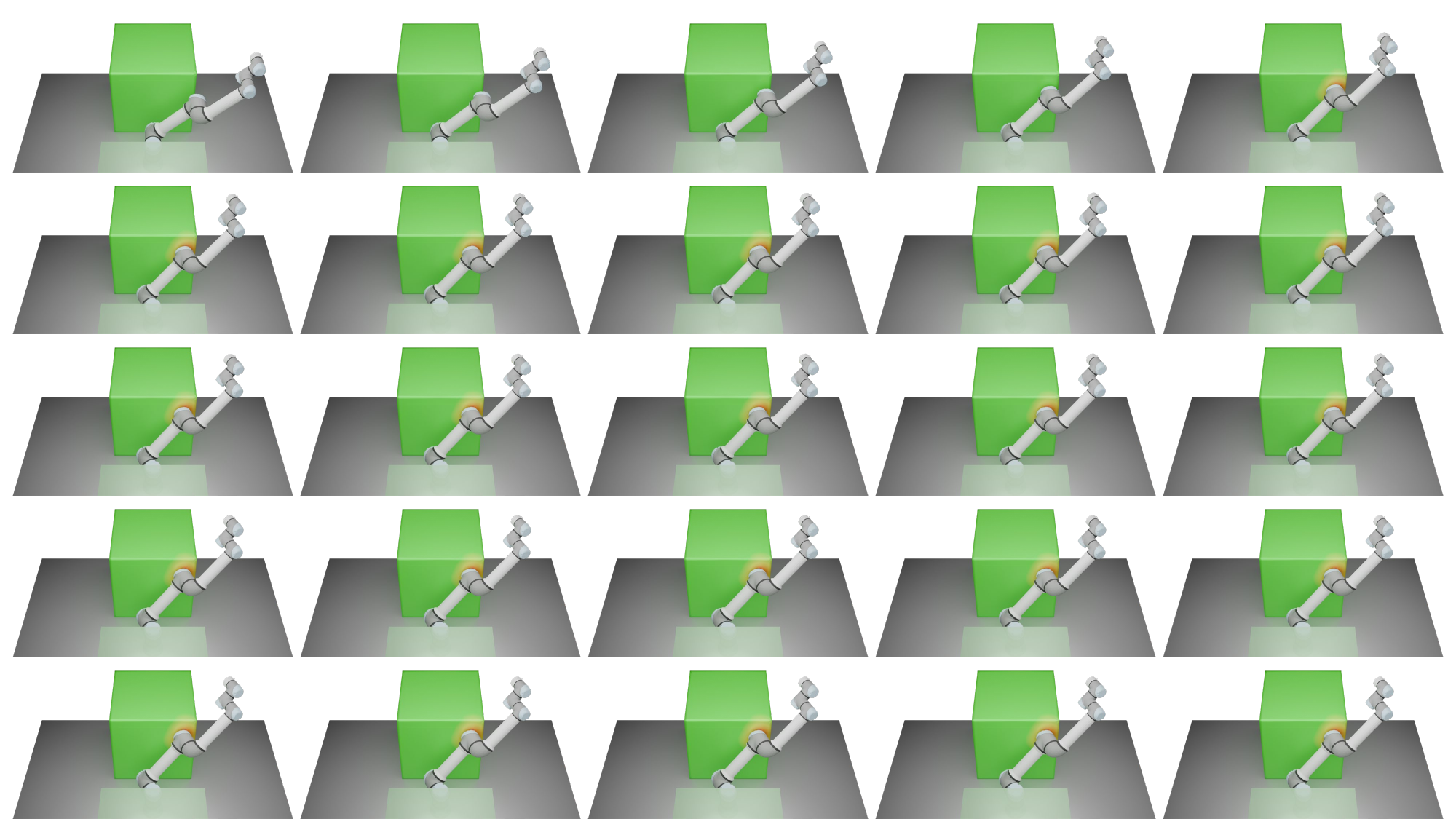}
    \caption{\textbf{Visualization of the use of the interpolation trajectory for robot control.}}
    \label{fig:visualization-of-using-interpolation-trajectory}
\end{figure*}

However, this elementary interpolation method may yield invalid angles and potential collisions, as illustrated in Figure \ref{fig:visualization-of-interpolation-trajectory}. If we were to rely on the trajectory generated by interpolation to instruct the robot's movements, it would face the risk of becoming ensnared at collision points, as depicted in Figure \ref{fig:visualization-of-using-interpolation-trajectory}. Thus, it is imperative to employ our SDF model to optimize the control points of the intermediate poses. By incorporating our SDF perception, we enhance the interpolation poses through the following formulation:
\begin{equation}
    \mathcal{L} = \gamma_1\mathcal{L}_\textrm{SDF} + \gamma_2\mathcal{L}_\textrm{TV}    
\end{equation}
where \(\gamma_1\) and \(\gamma_2\) represent scalar weights. The term \(\mathcal{L}_\textrm{TV}\) is detailed in Section \ref{sec:inverse_trajectory_from_opt} and is expressed as:
\begin{equation}
    \mathcal{L}_\textrm{TV} = \frac{1}{N-1} \sum_{i=1}^{N-1} \|\hat{\bm{\theta}}_{i+1} - \hat{\bm{\theta}}_i\|
\end{equation}

\begin{figure*}
    \centering
    \includegraphics[width=\textwidth]{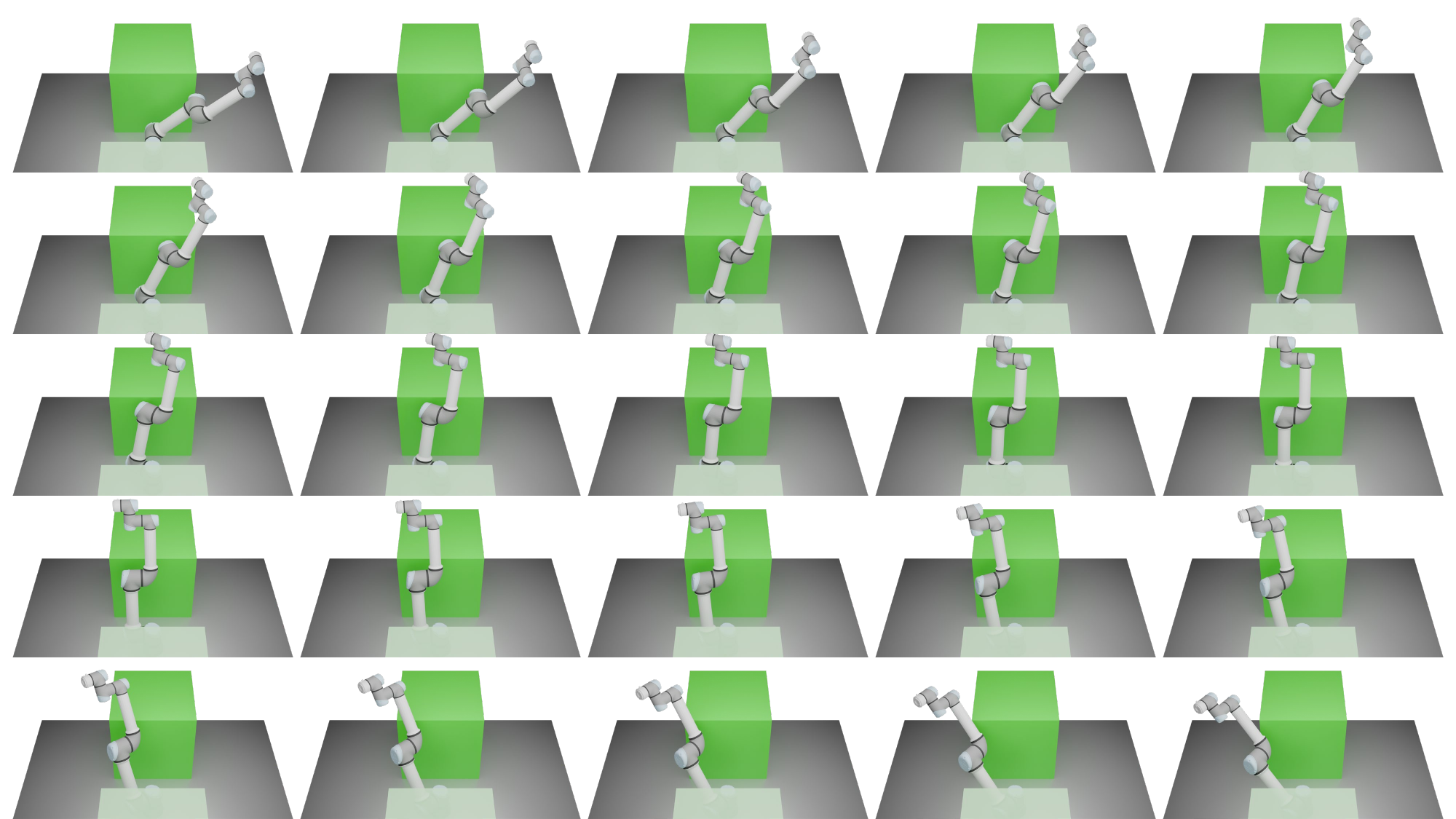}
    \caption{Visualization of our optimization trajectory derived from interpolation.}
    \label{fig:visualization-of-our-optimization-trajectory-from-interpolation}
\end{figure*}

\begin{figure*}
    \centering
    \includegraphics[width=\textwidth]{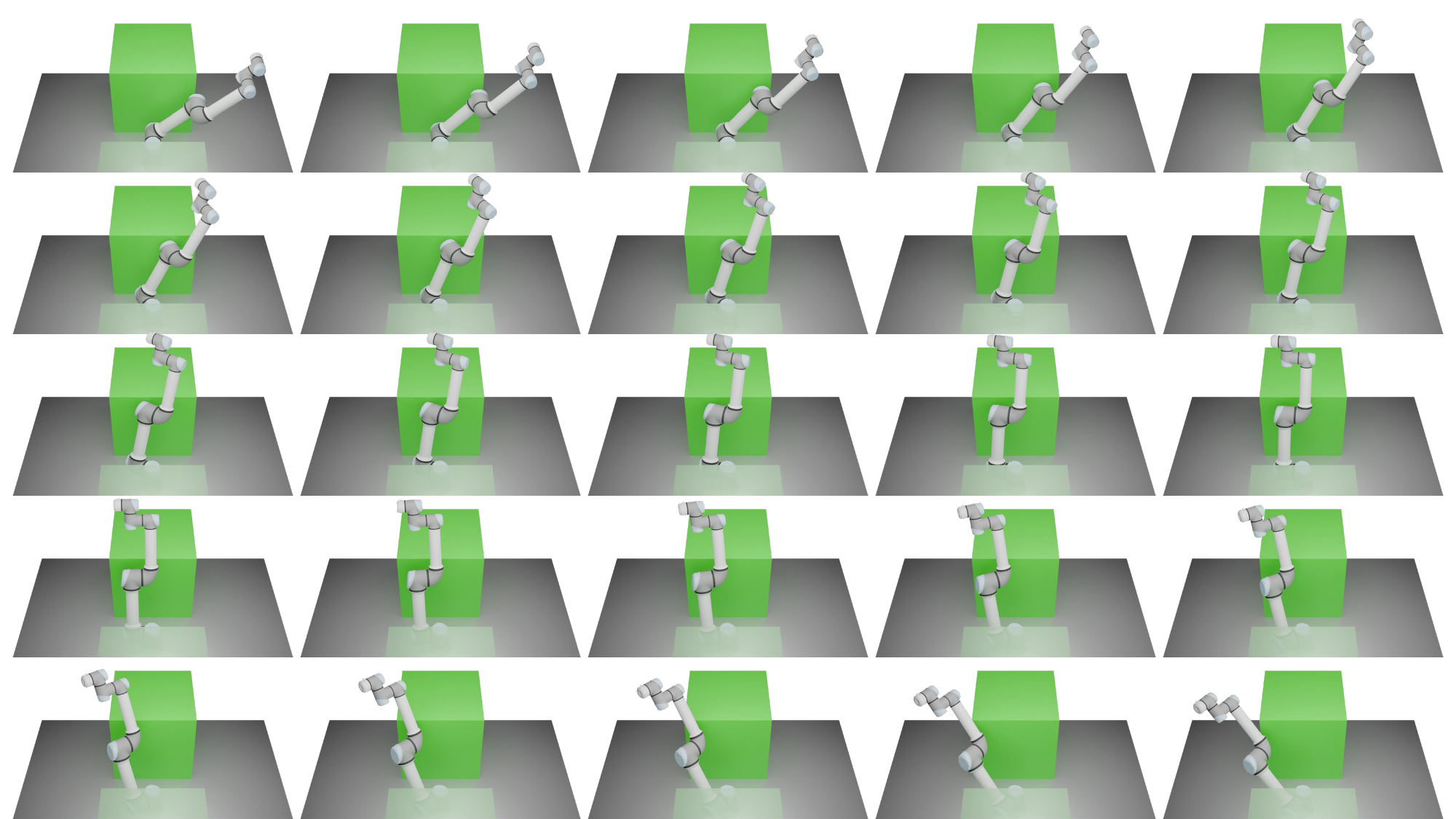}
    \caption{\textbf{Visualization of controlling the robot using our optimization trajectory from interpolation.}}
    \label{fig:visualization-of-using-our-optimization-trajectory-from-interpolation}
\end{figure*}

Utilizing our SDF model, we generate a collision-free trajectory to guide the robot's movements, as shown in Figure \ref{fig:visualization-of-our-optimization-trajectory-from-interpolation}. This trajectory is subsequently utilized to control the robot's motion, as illustrated in Figure \ref{fig:visualization-of-using-our-optimization-trajectory-from-interpolation}.

\subsection{Trajectory of Dr.~Robot}

\begin{figure}[h]
    \centering
    \includegraphics[width=0.9\linewidth]{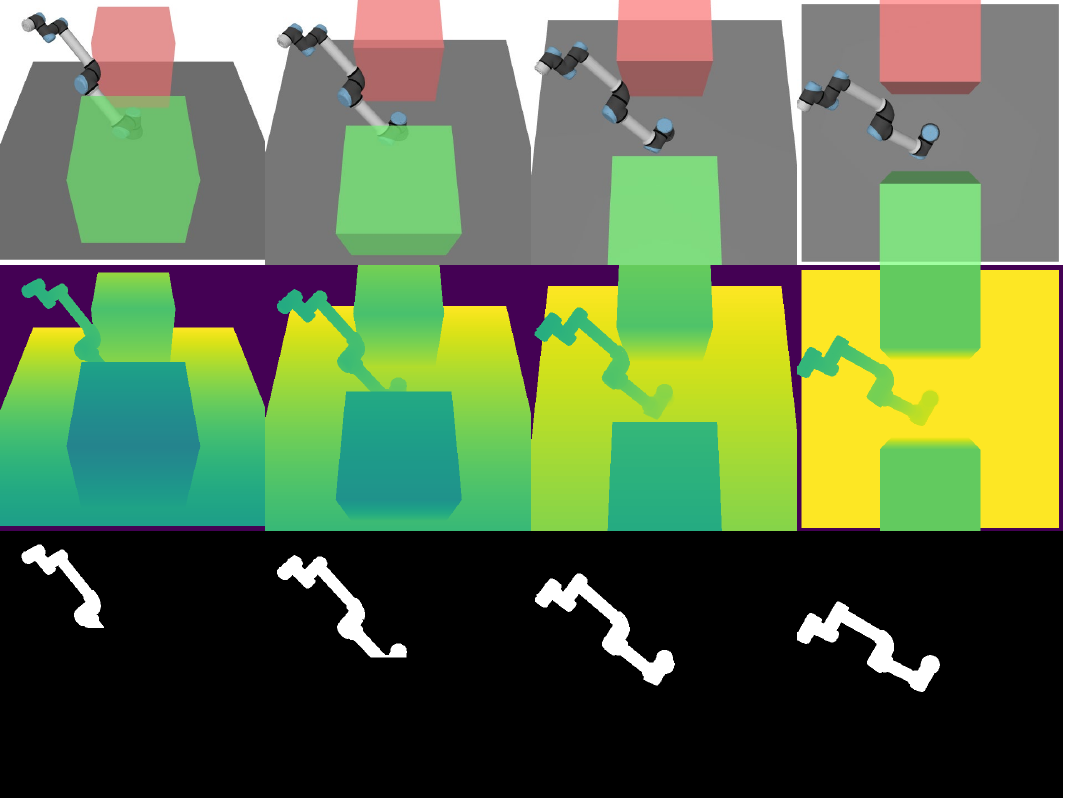}
    \captionof{figure}{\textbf{The Dataset for Trajectory Optimization from Dr.~Robot.} We use four different views as supervision, providing image, depth, and segmentation data for each perspective.}
    \label{fig:drrobot_trajectory_dataset}
\end{figure}

\begin{figure*}
    \centering
    \includegraphics[width=\textwidth]{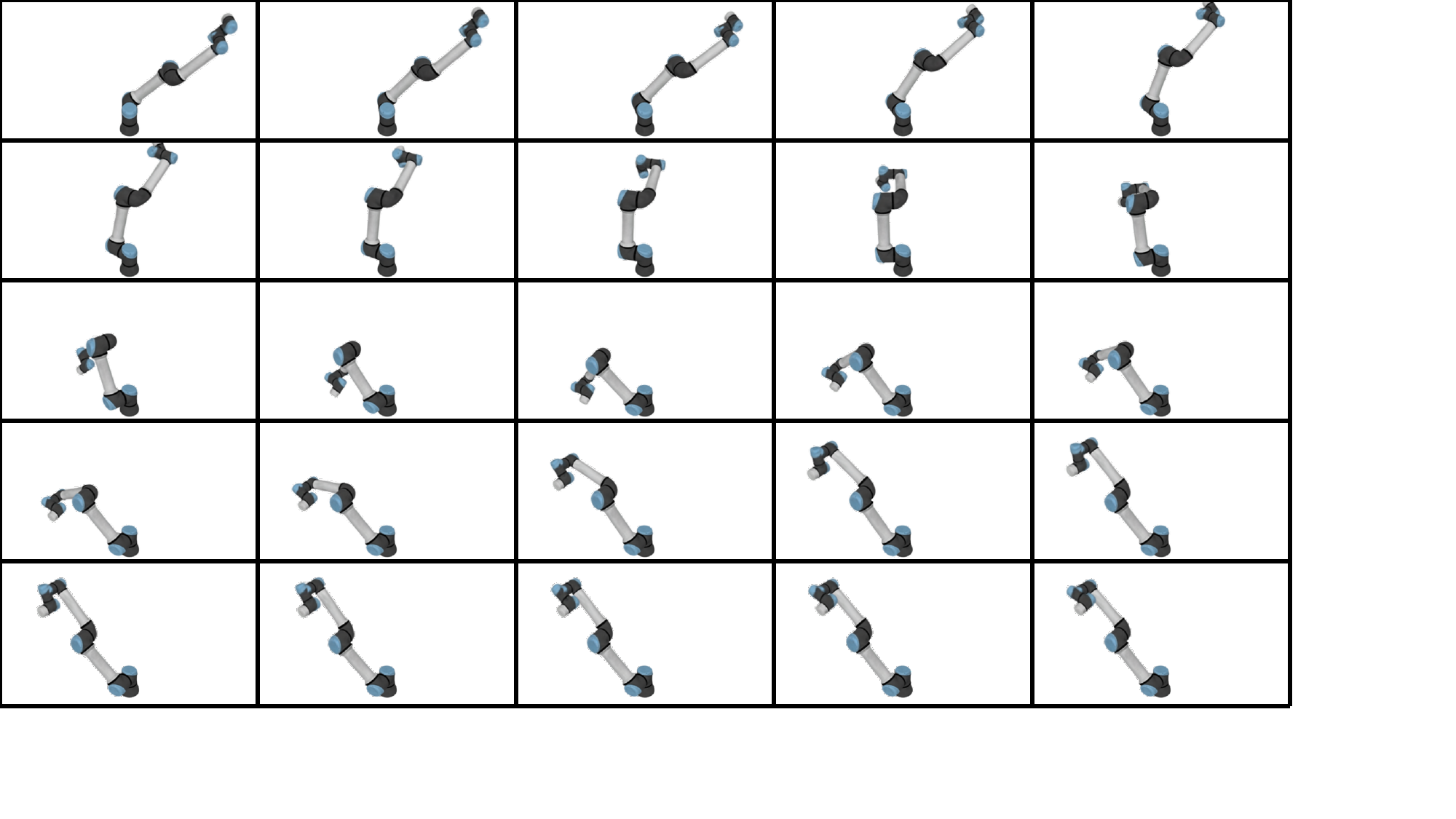}
    \caption{\textbf{Intermediate trajectory of Dr.~Robot's optimization process.}}
    \label{fig:drrobot_inverse_trajectory}
\end{figure*}

We render the target posture \(\bm\theta_\textrm{end}\) and utilize the intermediate optimized postures as control poses for Dr.~Robot. Four views of the target posture \(\bm\theta_\textrm{end}\) are presented in Figure \ref{fig:drrobot_trajectory_dataset}. Starting from the initial parameters \(\bm\theta_\textrm{start}\), the intermediate poses are optimized using the loss functions \(\mathcal{L}_\textrm{RGB}\) and \(\mathcal{L}_\textrm{D}\), as described in Section \ref{sec:inverse_action_from_collision_image}. The optimized intermediate postures are illustrated in Figure \ref{fig:drrobot_inverse_trajectory}.

\begin{figure*}
    \centering
    \includegraphics[width=\textwidth]{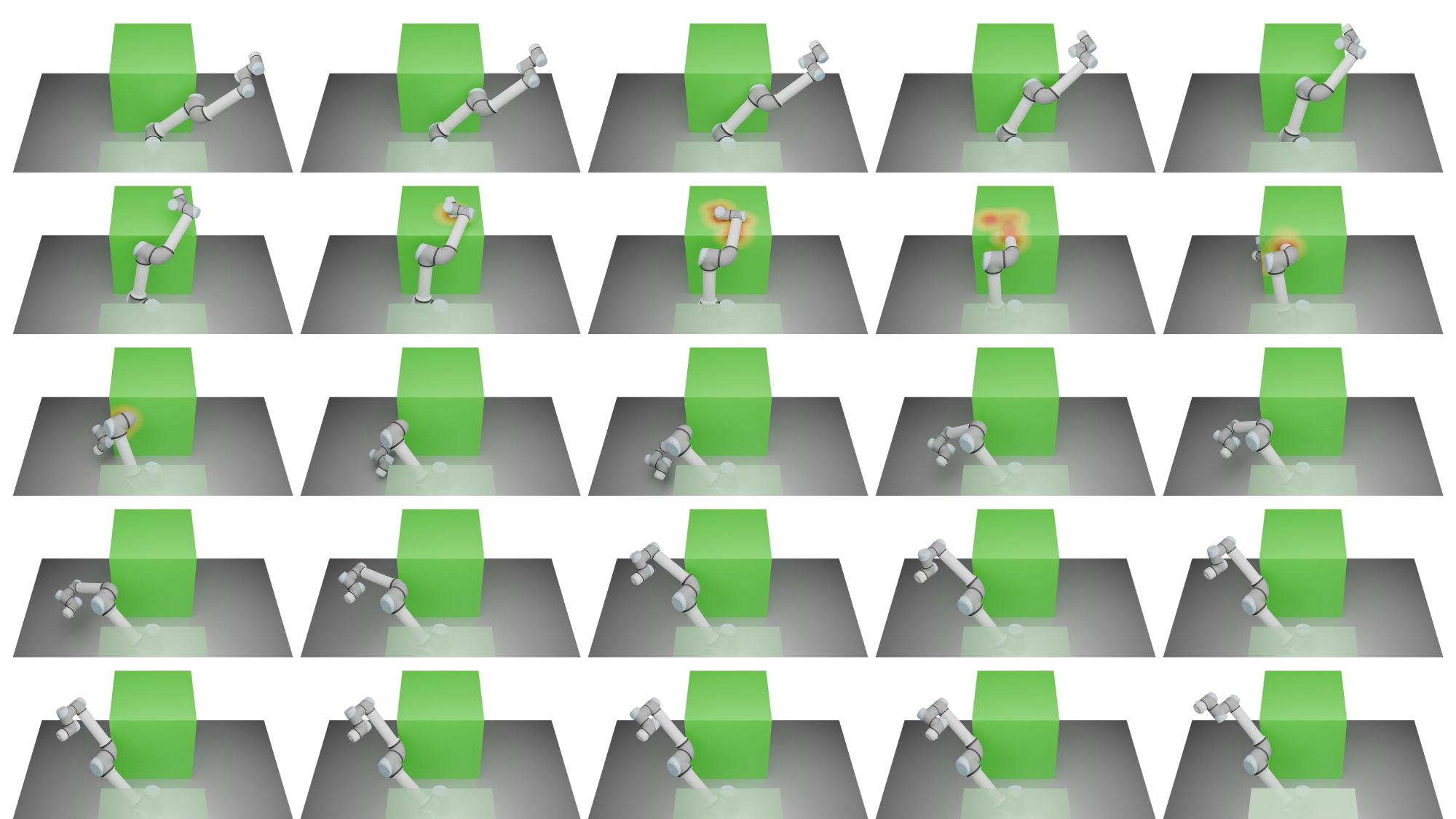}
    \caption{Visualization of the trajectory generated by Dr.~Robot.}
    \label{fig:drrobot_trajectory}
\end{figure*}

Directly using these postures for control can result in collisions, as shown in Figure \ref{fig:drrobot_trajectory}. To mitigate this issue, our method refines the pose optimization using the following formulation:
\begin{equation}
    \mathcal{L} = \gamma_1\mathcal{L}_\textrm{SDF} + \gamma_2\mathcal{L}_\textrm{TV} + \gamma_3\mathcal{L}_\textrm{SDF}
\end{equation}

\begin{figure*}
    \centering
    \includegraphics[width=\textwidth]{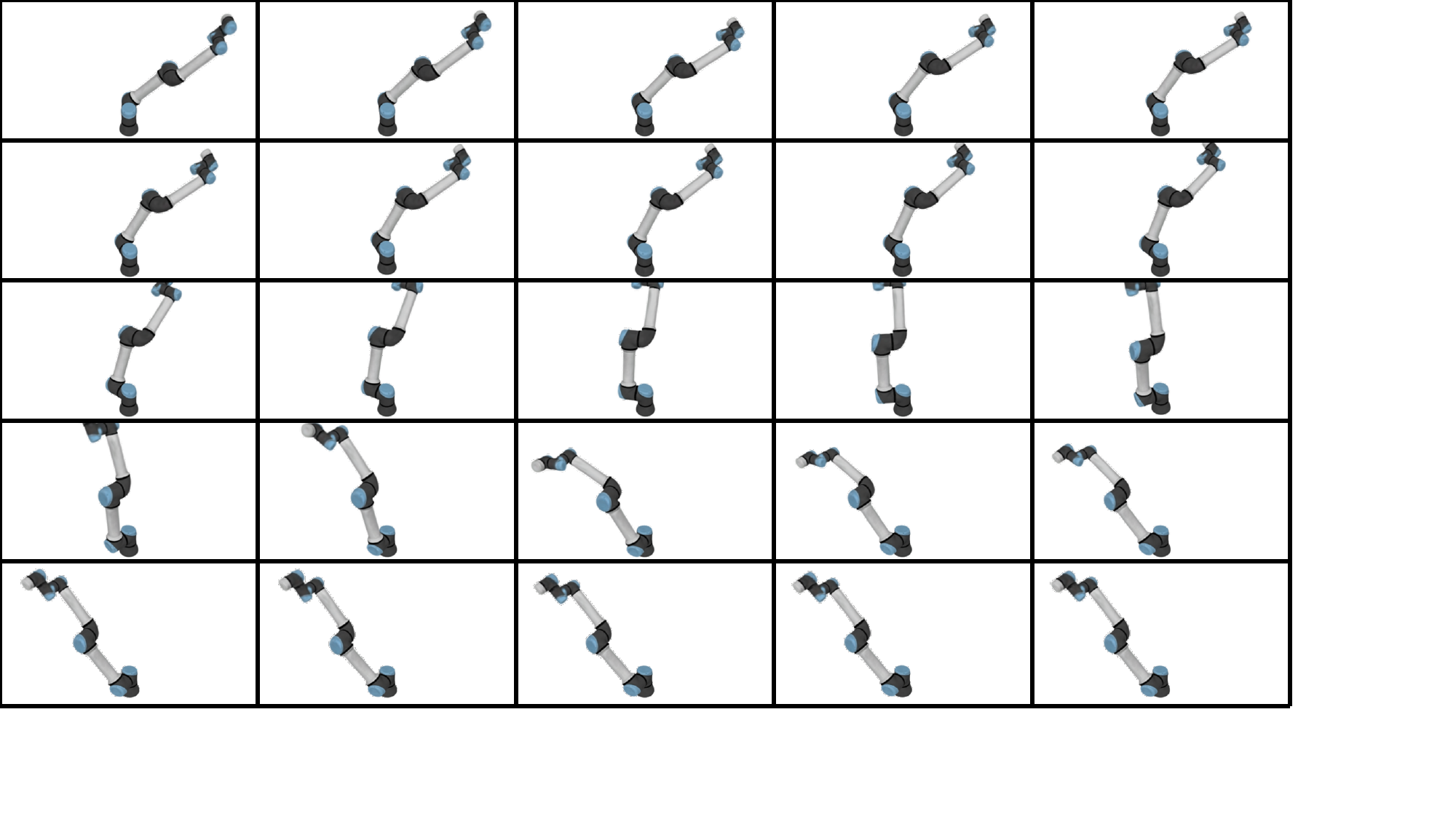}
    \caption{\textbf{Intermediate trajectory of Dr.~Robot's optimization utilizing our SDF model.}}
    \label{fig:ours_drrobot_inverse_trajectory}
\end{figure*}

\begin{figure*}
    \centering
    \includegraphics[width=\textwidth]{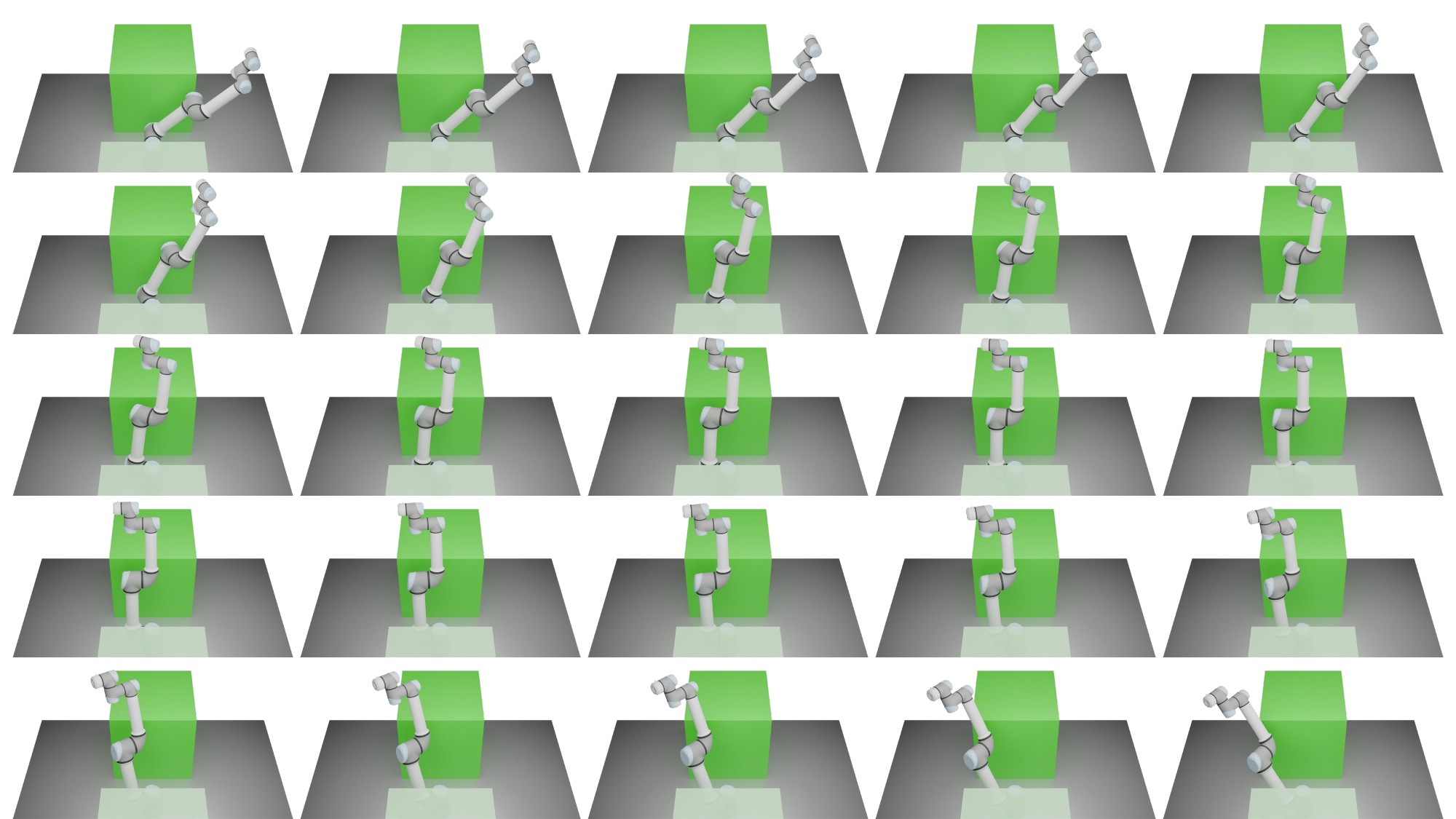}
    \captionof{figure}{\textbf{Visualization of Dr.~Robot's trajectory augmented by our SDF model.}}
    \label{fig:ours_drrobot_trajectory}
\end{figure*}

The results are shown in Figure \ref{fig:ours_drrobot_inverse_trajectory}. We further refine the optimization by employing \(\mathcal{L}_\textrm{SDF}\) and \(\mathcal{L}_\textrm{TV}\), leading to the control outcomes illustrated in Figure \ref{fig:ours_drrobot_trajectory}. These results clearly demonstrate that our optimization method significantly reduces collisions between the robot and its surrounding environment.

\subsection{Trajectory of Tangent}

Given the unique properties of the SDF, its value consistently increases along the gradient direction while remaining unchanged along tangential directions. However, with the multitude of possible tangential directions within the SDF, a crucial question emerges: which tangential direction should be chosen for gradient updates?

\begin{figure*}
    \centering
    \includegraphics[width=\textwidth]{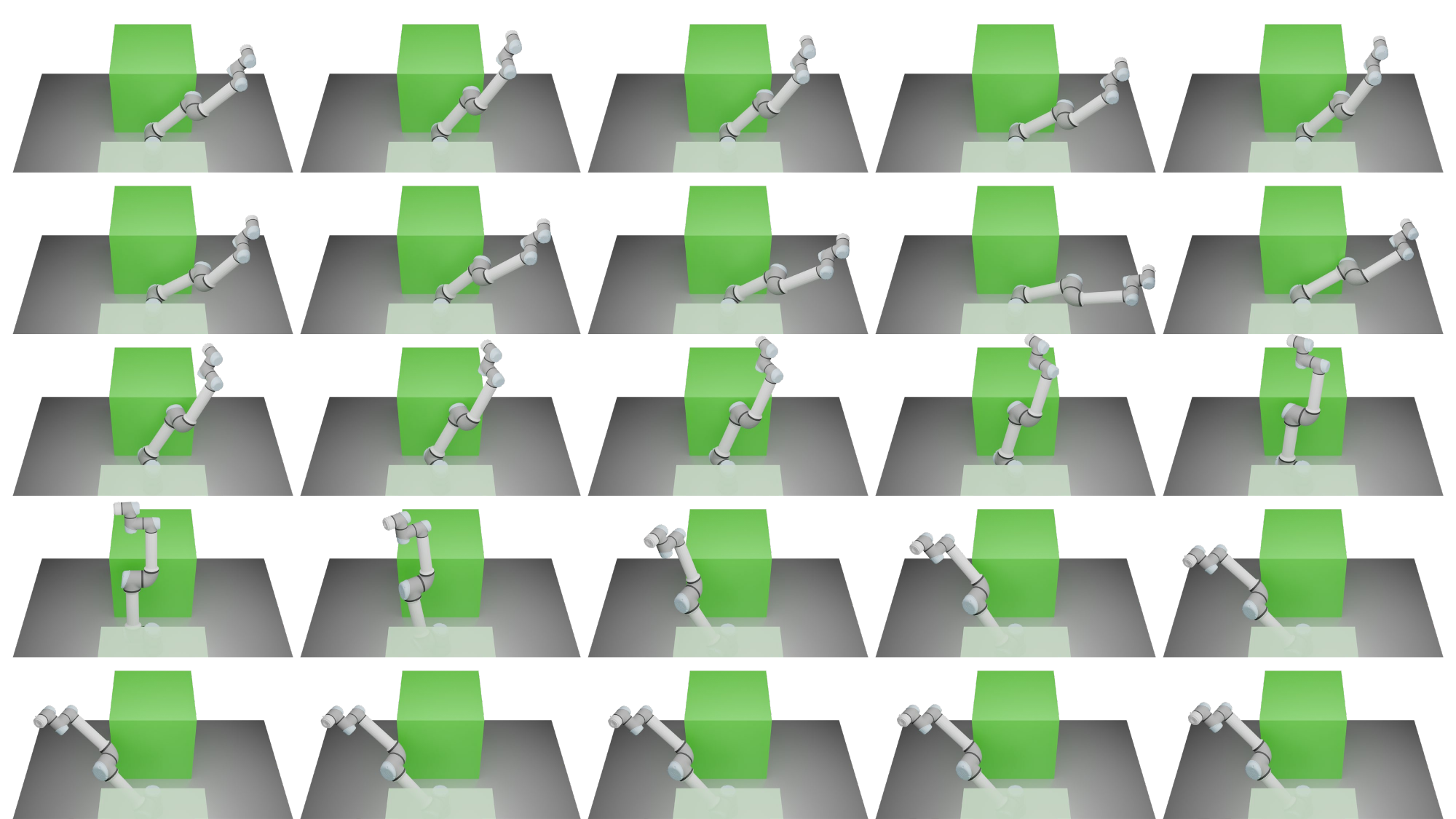}
    \captionof{figure}{\textbf{Visualization of our tangent trajectory}. Given that we employed only gradient descent without the inclusion of momentum for optimization, while assigning a substantial value to \( \delta \) to assist \( \bm\theta \) in circumventing local minima, this approach led to considerable jitter in the robot depicted in the image.}
    \label{fig:ours_tangent_trajectory}
\end{figure*}

To address this, we leverage additional available information to guide the gradient. By projecting this external information onto the tangential plane of the SDF gradient, we can define an appropriate update direction. We propose Algorithm \ref{alg:gradient-based-optimization-with-sdf-constraints}, which facilitates movement within a plane where the SDF value remains constant. This approach effectively enables the generation of a collision-free trajectory. The resulting outputs are depicted in Figure \ref{fig:ours_tangent_trajectory}.

\begin{algorithm}
\caption{Gradient-Based Optimization with SDF Constraints}
\label{alg:gradient-based-optimization-with-sdf-constraints}
\begin{algorithmic}[1]
\Require $\bm\theta_{\textrm{start}}$, $\bm\theta_{\textrm{end}}$, $\delta$, \textrm{maxIterations}, $\varepsilon_\textrm{tolerance}$
\Ensure Optimized $\bm\theta$

\State Initialize $\bm\theta \gets \bm\theta_{\textrm{start}}$
\For{$\text{iteration} = 1$ to \textrm{maxIterations}}
    \State Compute the SDF \(g(\bm\theta)\) and the gradient $\bm G_{\textrm{SDF}}$
    \State Normalize the $\bm G_{\textrm{SDF}}$: $\bm G_{\textrm{SDF}}^{\textrm{norm}} \gets \frac{\bm G_{\textrm{SDF}}}{\|\bm G_{\textrm{SDF}}\|}$
    \State Calculate the L1 distance to $\bm\theta_{\textrm{end}}$: $\mathcal{L}_{\bm\theta} \gets \left|\bm\theta-\bm\theta_{\textrm{end}}\right|$, \newline
           and compute the gradient of $\mathcal{L}_{\bm\theta}$: $\bm G_{\bm\theta}$
    \State Determine the projection of $\bm G_{\bm\theta}$ onto the tangent of 
            \hspace{3cm} $\bm G_{\textrm{SDF}}^{\textrm{norm}}$: 
    $\bm D \gets \bm G_{\bm\theta} - (\bm G_{\textrm{SDF}}^{\textrm{norm}} \cdot \bm G_{\bm\theta}) \bm G_{\textrm{SDF}}^{\textrm{norm}}$
    \State Update $\bm\theta$: $\bm\theta \gets \bm\theta - \delta \bm D$
    \If{$\mathcal{L}_{\bm\theta} < \varepsilon_\textrm{tolerance}$}
        \State \textbf{break}
    \EndIf
\EndFor
\end{algorithmic}
\end{algorithm}

\section{Comparison with Differentiable Simulator}

We frame Prof.~Robot as a lightweight plug-in that can be seamlessly integrated into other skill-learning pipelines. 
In specific, Prof.~Robot has a compact JIT model size of only 3.1 MB, as opposed to differentiable simulation, which requires a storage of nearly 4.7 GB and additional robot assets of nearly 22.1 MB. 
When optimizing poses, our model and the PyTorch environment require just 712 MiB of GPU memory. In comparison, Brax consumes 11,445 MiB, making our approach approximately 16 times more memory-efficient.
When optimizing poses, our approach approximately 16 times more memory-efficient.
Additionally, while differentiable simulators are capable of backpropagating gradients from collisions to actions, they come with significant limitations, such as computational overhead and inaccurate gradient estimation \cite{newbury2024review}.
Moreover, many simulators, such as MuJoCo and Isaac Gym, are not differentiable, and our Prof.~Robot is compatible with these simulators.

\section{Train with Alternative Regularizers}
Although optimizations involving Eikonal equations are known to be unstable, recent work by \cite{yang2023steik} proposes a stable Eikonal loss (StEik) formulation.
While incorporating StEik into our framework yields a modest 0.29\% improvement in classification accuracy, we observe that its effectiveness is constrained by fundamental factors: 
StEik primarily targets optimization of complex geometric configurations, whereas our pose optimization task does not strictly require full adherence to the Eikonal property. 
Similarly, imposing Lipschitz constraints \cite{liu2022learning} improves classification accuracy by 0.35\% through gradient stabilization, but introduces unintended consequences. 
This regularization relaxes the critical gradient normalization condition, thereby compromising essential SDF properties: 
(1) the geometric interpretation of SDF values as exact distance measurements to zero-level surfaces.
(2) the directional significance of SDF gradients for surface orientation. 
Such properties prove crucial for downstream tasks like pose interpolation \cite{tiwari2022pose}.

\section{Manipulation with Dr.~Robot}

To showcase the efficacy of our method in robot control, we undertook a training regimen for Dr.~Robot, enabling it to learn and execute actions based on video sequences. The effectiveness of this learning was then verified through a series of simulation experiments. We created a multi-view video of a pick operation using the Mujoco engine and employed Dr.~Robot to optimize the actions required for each video frame. Subsequently, we assessed these optimized actions within the simulation environment, with a focus on analyzing their interaction with the desk to detect any collisions.


In the robot demonstration video, each frame corresponds to a control parameter \( \bm\theta_i \). Our objective is to infer these \( \bm\theta_i \) values from the video and subsequently use them to control the robot. While Dr.~Robot has the capability to manipulate a robot, no demonstrations of this functionality have been presented so far. We are confident that with high-quality video input, Dr.~Robot can effectively manipulate a real robot. 

\begin{figure}[h]
    \centering
    \includegraphics[width=0.9\linewidth]{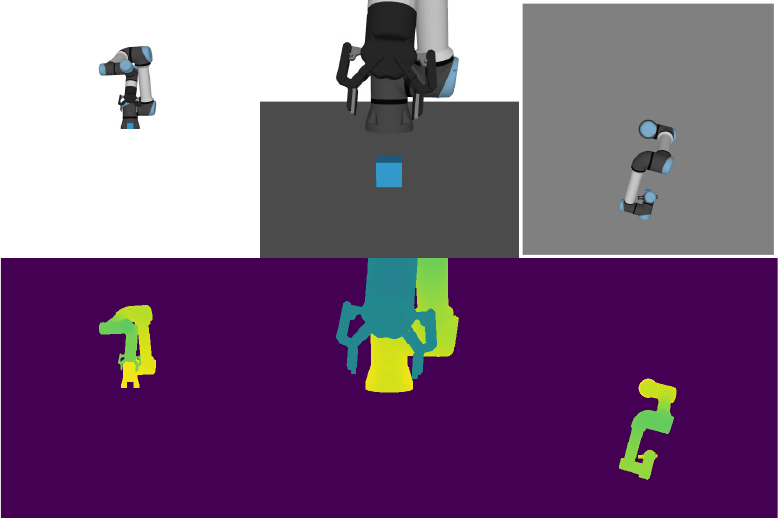}
    \captionof{figure}{\textbf{The Dataset for Robot Manipulation}. Our dataset includes three perspectives captured across 160 frames. For supervision, we provide images alongside their corresponding depth data of the robot. To address the singularity of actions, we ensure multi-view consistency by utilizing three distinct perspectives. Additionally, to supervise fine-grained manipulation, we use a telephoto camera to capture close-up images of the robot's gripper.}
    \label{fig:learn_from_video_visibe}
\end{figure}

For the sake of brevity, we do not focus on the quality of the survival model; instead, we directly learn the robot's actions from the demonstration video. Using the Mujoco engine, we generate a robot manipulation video, as shown in Figure \ref{fig:learn_from_video_visibe}. To ensure precise manipulation data, we adopt a course-to-fine strategy. Finally, we incorporate our collision-aware module to prevent any incidents of collision during robot operation.

\begin{figure*}
    \centering
    \includegraphics[width=\textwidth]{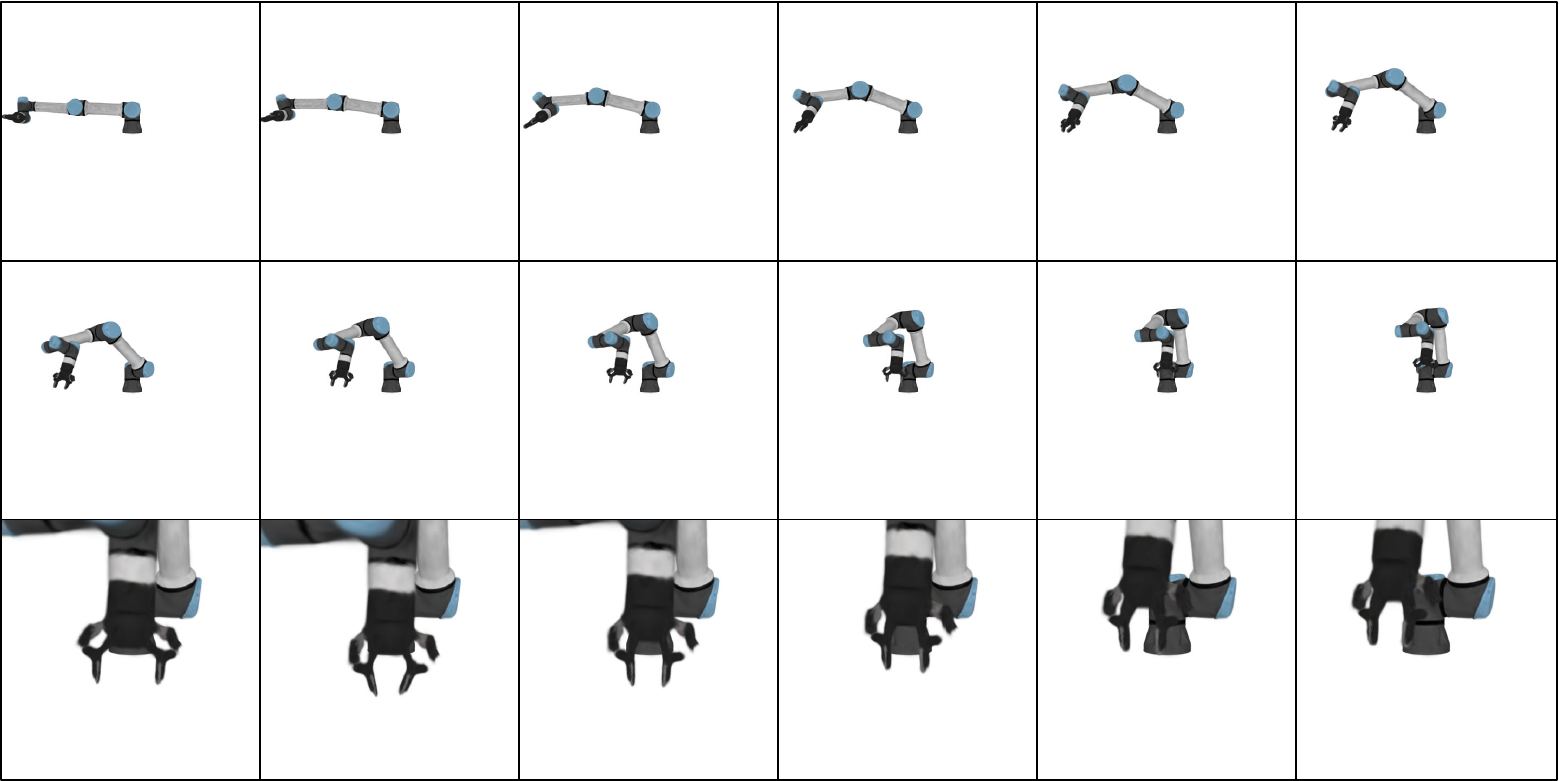}
    \captionof{figure}{\textbf{Rendering results of robot control utilizing Dr.~Robot.} The first two rows display operational images from a singular perspective, rendered by Dr.~Robot, while the final row illustrates the gripper being precisely manipulated by Dr.~Robot. Upon learning the parameter \( \bm\theta \), Dr.~Robot proceeds to control the robot.}
    \label{fig:learn_from_video_inverse_action}
\end{figure*}

During the course stage, we utilize gradient accumulation, leveraging Dr.~Robot to propagate the gradients from the three views before updating the parameters. In this phase, we adopt a larger learning rate and recursively optimize the parameters of the robot for each frame, performing five parameter updates before retaining the optimization results. In the fine stage, we still employ the method of gradient accumulation, but with a smaller learning rate and incorporate the TV loss mentioned in Section \ref{sec:inverse_trajectory_from_opt} to mitigate jitter during the optimization process, conducting ten parameter updates. The results are presented in Figure \ref{fig:learn_from_video_inverse_action}.

\begin{figure*}
    \centering
    \includegraphics[width=\textwidth]{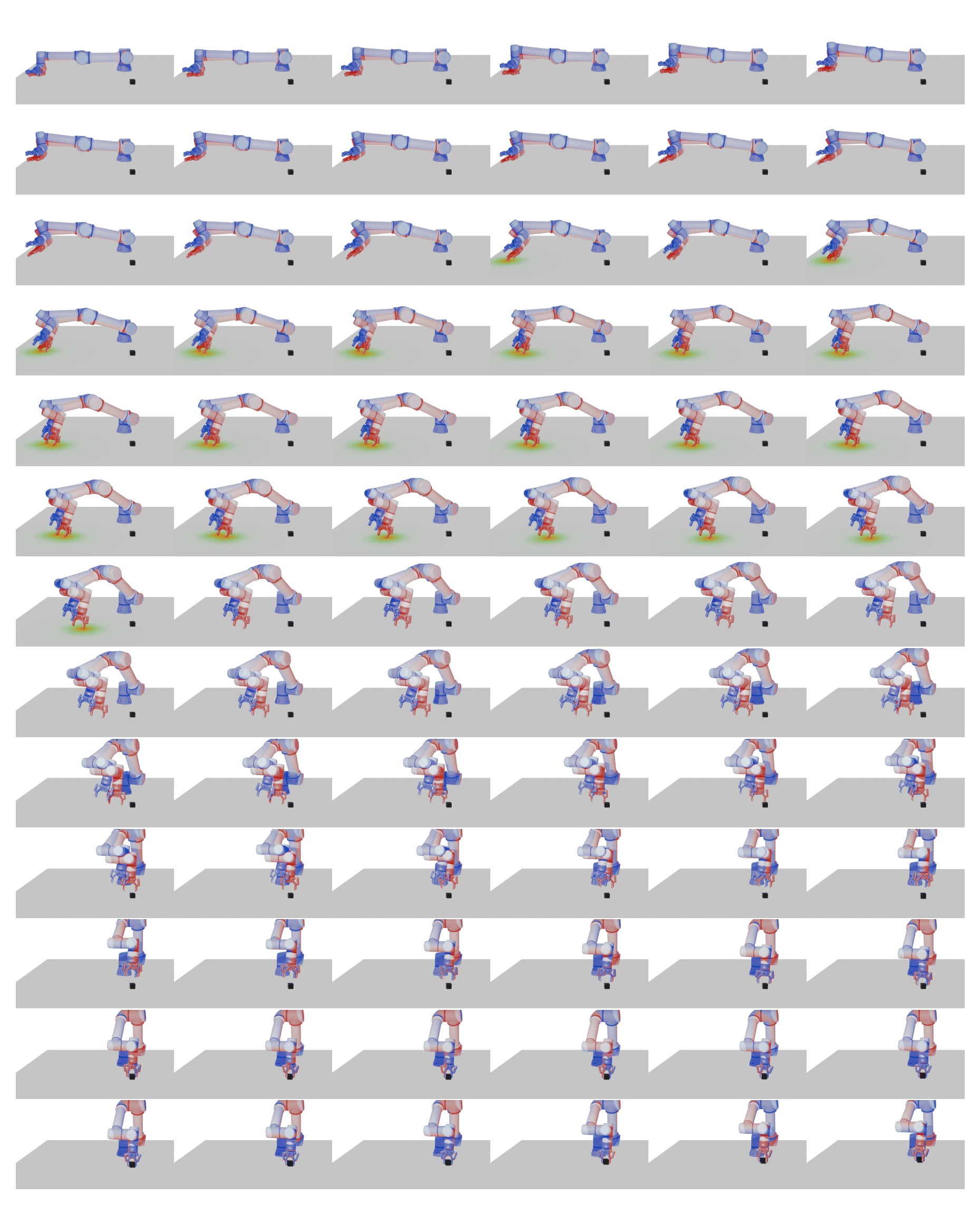}
    \captionof{figure}{\textbf{Results of robot control employing the parameters \( \bm\theta \) learned by Dr.~Robot}. The {\color{red} \textbf{red}} robot arm in the image denotes direct control based on the original parameters learned by Dr.~Robot, while the 
    {\color{blue} \textbf{blue}} robot arm represents control following optimization through our SDF model. To highlight potential collisions, we elevated the plane of the initial 60 views by 0.005 m solely during the rendering process, with collision points marked in {\color{red} \textbf{red}} on the {\color{gray} \textbf{gray}} desktop.}
    \label{fig:control-robot-by-drrobot}
\end{figure*}

However, in our demonstration scenario, directly utilizing Dr.~Robot for robot control can result in collisions. To circumvent this issue, we optimize the parameters \( \bm\theta \) estimated by Dr.~Robot using our SDF model. We categorize the robot's movements into two types: movement, which typically does not emphasize the posture of the end effector but focuses on avoiding collisions during motion, and operations, which prioritize the end effector's posture. Consequently, we classify the parameters \( \bm\theta \) inferred by Dr.~Robot into these two categories. For the movement parameters, we apply SDF loss for supervision, while for the operation parameters, we supervise based on the posture of the end effector. Throughout the optimization process, we also employ TV loss to ensure smoother postures and to prevent excessive jitter. Our evaluation primarily focuses on collision avoidance between the robot arm and the desktop during movement, as illustrated in Figure \ref{fig:control-robot-by-drrobot}. The results demonstrate that our method effectively mitigates robot collisions.

\end{document}